%% file: main.tex
\documentclass{article} 
\usepackage{main,times}

\input{utils/math_commands}

\usepackage{hyperref}
\usepackage{url}

\usepackage{multirow}
\usepackage{booktabs}
\usepackage[table]{xcolor} 
\usepackage{adjustbox}        
\usepackage{array}            
\usepackage{xcolor}
\usepackage{makecell}
\usepackage{wrapfig}
\usepackage{graphicx}
\usepackage{amssymb} 

\definecolor{streamnavy}{HTML}{172A63}
\definecolor{erasepurple}{HTML}{6B2FEA}
\definecolor{eraseviolet}{HTML}{6550E8}
\definecolor{eraseblue}{HTML}{387FE5}
\definecolor{erasecyan}{HTML}{28B9E8}

\NewDocumentCommand{\inlineimage}{O{0.5} m}{%
  \raisebox{-0.25\baselineskip}{\includegraphics[height=#1\baselineskip]{#2}}\hspace{-1pt}
}

\newcommand{\method}{StreamErase}

\title{\method: Towards Ultra-Fast Streaming Video Erasing}

\iclrfinalcopy

\author{%
  Zhuobai Dong$^{1}$, Xiaobing Tu$^{2}$, Xiantao Zhang$^{2}$, Jinkui Ren$^{2}$, \\
  \textbf{Yinggui Wang}$^{2}$, \textbf{Yue Ma}$^{3}$, \textbf{Linfeng Zhang}$^{4}$\thanks{Corresponding author.} \\
  $^{1}$Wuhan University \quad $^{2}$Alibaba Group \\
  $^{3}$The Hong Kong University of Science and Technology \quad $^{4}$Shanghai Jiao Tong University
}

\begin{document}


\pagestyle{empty} 
\maketitle
\thispagestyle{empty} 

\input{figure/tex/teaser}

\input{section/1_abstract}

\input{section/2_intro}

\input{section/3_related}

\input{section/4_method}

\input{section/5_exp}

\input{section/6_conclusion}

\input{section/9_ethics}

\bibliography{main}
\bibliographystyle{main}

\appendix
\input{section/8_appendix}

\end{document}

%% file: utils/math_commands.tex
\usepackage{amsmath,amsfonts,bm}

\def\eqref#1{equation~\ref{#1}}

\def\1{\bm{1}}

\DeclareMathAlphabet{\mathsfit}{\encodingdefault}{\sfdefault}{m}{sl}
\SetMathAlphabet{\mathsfit}{bold}{\encodingdefault}{\sfdefault}{bx}{n}



%% file: figure/tex/teaser.tex
\begin{figure*}[h]
    \vspace{-0.5cm}
    \centering
    \includegraphics[width=\textwidth]{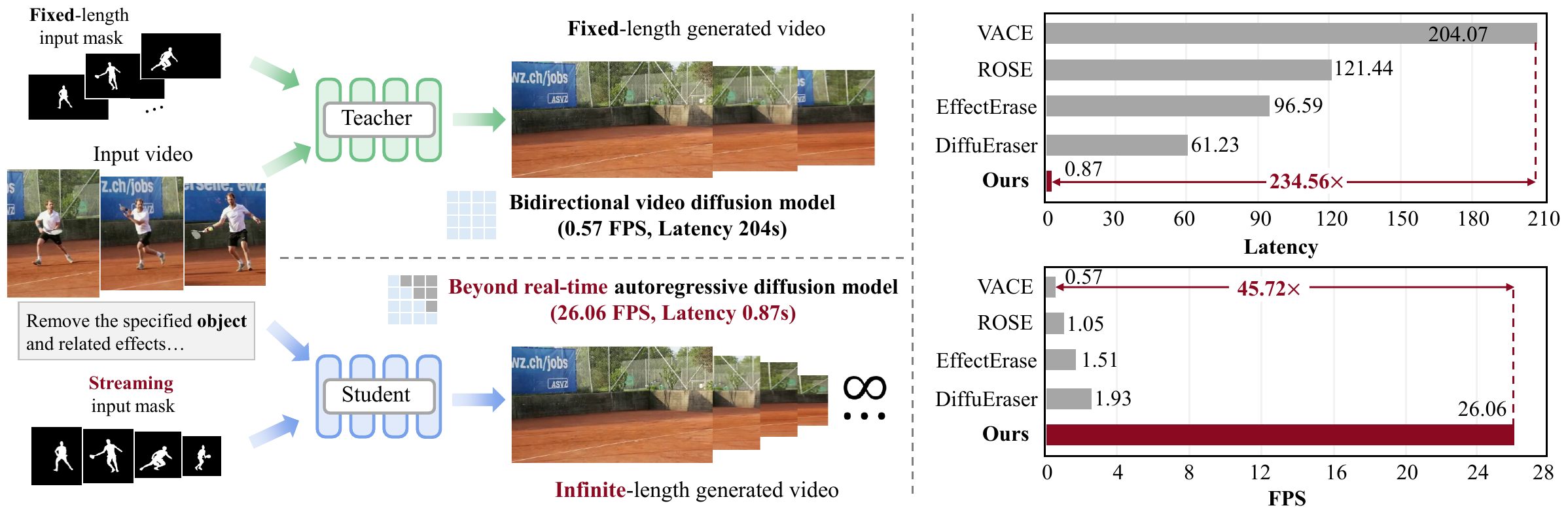}
    \vspace{-0.8cm}
    \caption{Prior video diffusion erasing models typically operate offline to generate fixed-length sequences in parallel (top left). In contrast, our \method{} enables streaming long-video generation (bottom left), supporting real-time erasing at 26.06 FPS with 0.87s latency (right).
    }
    \label{teaser}
\end{figure*}

%% file: section/1_abstract.tex
\begin{abstract}
Current video object removal methods suffer from prohibitive latency (minutes per video) and non-causal processing that prevents real-time interaction. 
We present \method, enabling sub-second latency with up to 26 FPS streaming generation on a single GPU. 
Our approach first augments a text-to-video model with mask input to generate high-quality videos that erase target objects and synthesize coherent backgrounds, but does not perform
inference on the fly. 
As such, we progressively transfer the erasing capability from a powerful bidirectional foundation model to an efficient unidirectional streaming eraser, enabling long-horizon erasing without sacrificing visual fidelity.
Several key challenges arise when generating videos over long, potentially infinite horizons: (1) bridging the domain gap from training on finite length and extrapolating to long horizons, (2) sustaining high quality by preventing error accumulation, and (3) maintaining fast inference, without increasing computational cost as the context grows.
To address these challenges, we carefully designed sliding-window causal attention with attention sinks.
By incorporating self-rollout with attention sinks and KV cache rolling during training, we simulate inference-time extrapolations with a fixed context window, enabling constant-speed generation of arbitrarily long videos.
Finally, we establish a dedicated benchmark, \textit{LongErase}, for streaming long-video object removal.
Our model achieves strong performance compared to existing video object erasing models and generalizes well to real-world video scenarios, uniquely enabling infinite-length streaming. 
With \method, users can input a mask and see the erasing results unfold in real-time, delivering a truly interactive experience.
HomePage: \url{https://zhuobaidong.github.io/StreamErase/}
\end{abstract}

%% file: section/2_intro.tex
\section{Introduction}

The ultimate goal of video object erasing is to empower creators and systems to seamlessly remove unwanted dynamic content from video streams in real time while preserving visual realism. Although recent video diffusion models~\citep{li2025diffueraser,bian2025videopainter,zi2025minimaxremovertamingbadnoise,vace,miao2025rose,fu2026effecterase} have made impressive strides toward this goal, the experience remains largely confined to offline, batch-processing workflows.

The promise of real-time video erasing is currently hindered by several fundamental constraints. 
First, generation is too slow for interactive deployment. For example, synthesizing a 5-second video clip with ROSE~\citep{miao2025rose} takes 2 minutes, trapping users in frustrating "render-and-wait" delay cycles. 
Second, existing paradigms are inherently non-causal, since diffusion models process the entire sequence in parallel with bidirectional attention. A user cannot see any partial results until the entire erase is complete. 
Finally, the inability to generate more than a few seconds of video severely limits the scope for any meaningful or extended creative expression. 
Together, these constraints (slow, non-causal, and short-duration generation) undermine the potential for a truly interactive creative experience.

To overcome these challenges, we introduce \method, a streaming video erasing framework that progressively transfers the erasing capability of an offline bidirectional diffusion model into an efficient autoregressive generator for real-time interactive erasing.

Specifically, we first fine-tune a pretrained diffusion model with full bidirectional attention, obtaining in a powerful bidirectional video eraser that serves as the teacher for subsequent distillation. We then replace bidirectional attention with causal attention to construct an autoregressive diffusion model, and apply teacher forcing followed by ODE distillation to obtain a causal video eraser requiring four denoising steps. Finally, we perform Distribution Matching Distillation (DMD), using the bidirectional video eraser as the teacher and the causal video eraser as the student, enabling the streaming model to inherit the high-fidelity erasing capability of the offline model while further compressing generation to a single denoising step for real-time inference.

Causal generation, however, is prone to error accumulation and drift over long horizons.
Prior work in large language models has shown that initial tokens play a crucial role in stable streaming generation~\citep{xiao2023efficient}.
This insight drives our attention sinking mechanism with rolling KV caches, which we incorporate directly into training to simulate inference extrapolation distributions, ensuring stable, indefinite-length generation at constant latency through fixed context windows.

\method{} achieves 26.06 FPS with sub-second latency on a single H200 GPU. Extensive experiments and ablations demonstrate that \method{} achieves competitive visual and temporal quality across diverse erasing scenarios, while being more than 46$\times$ faster than recent methods.
Furthermore, \method{} maintains stable visual and temporal quality when extended from short 5-second clips to substantially longer 30-second videos.
\method{} transforms video generation from a passive waiting experience into an active creative process, where users can continuously interact with and guide the generation in real-time.







%% file: section/3_related.tex
\section{Related Work}

\subsection{Video Inpainting}
Video inpainting aims to reconstruct missing regions specified by a sequence of masks.
Early video inpainting methods primarily used 3D CNNs~\citep{freeformvideoinpainting,videoinpaintingbyjointlylearning,proposalbasedvideocompletion} to model spatial-temporal features, but their limited receptive fields hindered long-range propagation. 
Subsequently, optical flow~\citep{kim2019deepvideoinpainting,li2020shortlongterm,zou2021progressive} and homography~\cite{lee2019copypaste,cai2022devit} were introduced to guide pixel propagation. 
To improve efficiency and accuracy, \citep{zhou2023propainter} proposed ProPainter, combining optical flow and attention mechanisms. 
Recently, with the rise of diffusion models\citep{ho2020ddpm}, diffusion-based video inpainting has emerged~\citep{shi2024bivdiff,zi2025cococo}. \citep{li2025diffueraser} proposed DiffuEraser, extending the image inpainting model BrushNet\citep{ju2024brushnet} to videos via a two-stage training scheme.
Despite these advances, existing inpainting models often overlook object effects, resulting in incomplete or visually inconsistent object removal.

\textbf{Video Object Erasing} is a specialized form of inpainting that requires precise modeling of object-induced visual effects to achieve realistic results.
Early methods ~\citep{yu2025omnipaint, wei2025omnieraser,liu2025erase} attempt to adapt image erasing models to video by processing frames individually, but struggle to maintain temporal consistency in dynamic scenes.
Minmax-Remover~\citep{zi2025minimaxremovertamingbadnoise} simplifies a pre-trained video generator by removing text inputs and cross-attention layers and introducing a minimax objective.
ROSE~\cite{miao2025rose} introduces a synthesized dataset comprising multiple environments, along with a side-effect mask predictor.
EffectErase~\citep{fu2026effecterase}  further provides a large-scale dataset of paired real and synthetic videos and proposes an effect-aware removal framework based on reciprocal learning.
Despite their impressive quality, these methods are fundamentally limited to offline processing because they rely on diffusion models with full bidirectional attention. This constraint prevents their use in real-time applications.

\subsection{Streaming Autoregressive Video Models.}
Streaming autoregressive generation effectively overcomes the latency bottlenecks of offline models. 
Foundational works like Diffusion Forcing~\citep{chen2024diffusion} and StreamDiffusion~\citep{kodaira2025streamdiffusion} reformulates denoising as block-wise sequential processing, enabling causal decoding. 
Follow-up works~\citep{yin2024slow,chen2025skyreels,teng2025magi,guo2025long,huang2026self,lin2025autoregressive} combined causal attention, KV-cache, and distillation to compress multi-step diffusion into a few denoising steps, enabling efficient online generation.
However, these approaches either exhibit severe color drifts beyond the training horizon or require complex long-video finetuning, which poses challenges for controllable video generation.

\subsection{Diffusion Distillation}

Accelerating reverse diffusion~\citep{song2023consistencymodels,wang2023videolcm,salimans2022progressive,song2023improved} is crucial for real-time performance. 
Early one-step distillation efforts focused on the image domain, utilizing techniques like Rectified Flow (InstaFlow~\citep{liu2023instaflow}), target score distillation (TSD-SR~\citep{Dong_2025_CVPR}), and Distribution Matching Distillation~\citep{yin2024dmd,yin2024dmd2}, alongside progressive adversarial distillation (SDXL-Lightning~\citep{lin2024sdxl}). 
These advancements have naturally extended to the video domain, enabling efficient one-step generation (AAPT~\citep{lin2025diffusion}), restoration (SeedVR2~\citep{wang2025seedvr2}), and super-resolution (DOVE~\citep{chen2025doveefficientonestepdiffusion}).
FlashVSR~\citep{zhuang2025flashvsr} achieves real-time streaming video super-resolution via a three-stage distillation pipeline designed to progressively accelerate latent rendering. Similarly, PersonaLive~\citep{li2025personalive} leverages appearance distillation for live portrait animation. 
However, these tasks involve low-level pixel mapping or structurally constrained regions. Directly transferring these schemes to general video erasing degrades high-fidelity details.

%% file: section/4_method.tex
\section{Method}
\subsection{\method}
\label{sec:meth:method}
\paragraph{Overview.}
In this section, we elaborate the model architecture of \method{} for conducting video object erasing, as illustrated in Fig.~\ref{method}. In brief, \method{} is implemented as an inpainting model continued from the foundation video generative models (the Wan2.1 model~\citep{wan2025wan} in this paper). Following the
general architecture in diffusion-based inpainting models~\citep{miao2025rose}, we extend the model input with the original input video together with object masks. 

\input{figure/tex/method1}

\noindent \textbf{Removal Formulation.}\quad
Video inputs are first encoded into the latent space using a pretrained VAE~\citep{kingma2013auto}.
The source video with objects $V^{o}$, the ground-truth background video $V^{b}$, and the mask ${M}$ are encoded into latent representations $\boldsymbol{x}^{o}$, $\boldsymbol{x}^b$, and $\boldsymbol{x}^{m}$, respectively.

To construct the noisy input $\boldsymbol{x}_{t}$ for diffusion training, a clean latent $\boldsymbol{x}$ obtained from the VAE is used. Random noise $\boldsymbol{z} \sim \mathcal{N}(0, I)$ is added through the forward process~\citep{esser2024scaling}:
\begin{equation}
\boldsymbol{x}_{t} = t\boldsymbol{x} + (1-t)\boldsymbol{z},
\end{equation}
where the timestep $t \in [0,1]$ is sampled from a logit-normal distribution. 
The denoising model $v_\theta$ is trained to predict the velocity $\boldsymbol{v} = \boldsymbol{x} - \boldsymbol{z}$ from the noisy latent $\boldsymbol{x}_{t}$, the timestep $t$, and the condition $\boldsymbol{c}$, with the objective defined as:
\begin{equation}
\mathcal{L}_\text{{diffusion}} = \mathbb{E}_{\boldsymbol{z}, \boldsymbol{x}, t, \boldsymbol{c}}\big\| v_\theta(\boldsymbol{x}_{t}, t, \boldsymbol{c}) - \boldsymbol{v} \big\|^2,
\end{equation}
where the condition $\boldsymbol{c} = [\boldsymbol{x}^o ; \boldsymbol{x}^m]$ guides the model to user-specified regions. Here $[\,;\,]$ denotes concatenation along the channel dimension.

\noindent \textbf{Multimodal Region Guidance.}\quad
To explicitly model spatiotemporal correlations between the affected areas and objects, we introduce a Region Guidance module.
Specifically, textual prompt tokens $\boldsymbol{e}^\text{{prompt}}$ are extracted from a language model~\citep{raffel2020exploring}, Simultaneously, visual foreground features $\boldsymbol{e}^{f}$ are obtained by feeding a cropped foreground patch from $V^{f} = V^{o} \odot M$ into a CLIP image encoder~\citep{radford2021learning}, where $\odot$ denotes element-wise multiplication.
To bridge the modality gap,  we employ a lightweight two-layer MLP $\mathcal{M}_{\psi}(\cdot)$ to map the CLIP embedding into the text token space.
The resulting foreground embedding $ \mathcal{M}_{\psi}(\boldsymbol{e}^{{f}}) $ then replaces the placeholder token ``object'' in $\boldsymbol{e}^{\text{prompt}}$, forming a region representation:
\begin{equation}
\boldsymbol{e}^\text{{prompt}}[\text{object}] \leftarrow \mathcal{M}_{\psi}(\boldsymbol{e}^{{f}}),
\end{equation}
which is injected into the backbone via cross-attention~\citep{vaswani2017attention} to guide the model in capturing spatiotemporal effect correlations between the object and its effects, enabling accurate localization of effect-related regions.

\noindent \textbf{Effect Erasing Loss.}\quad
Beyond the diffusion loss targeting reconstruct the regions of object and its side effect, we introduce an auxiliary Effect Erasing loss ($\mathcal{L}_\text{EE}$).
The core idea is to leverage the complementary information in the video pairs for training. 
When an object is removed from a scene, it often leaves behind subtle but semantically significant side effects, such as shadows, reflections, and occlusions.
To explicitly guide the model in attending to these regions, we compute a binary difference mask by comparing the original video $\mathbf{V}^{o} \in \mathbb{R}^{c \times f \times h \times w}$ with the background video $V^{b}$. The difference mask \( \mathbf{d}_0 \in \{0,1\}^{f \times h \times w} \) is defined as:
    \begin{equation}
        \mathbf{d}_0^{(t,h,w)} = 
        \begin{cases}
            1, & \text{if } \left\| {{\mathbf{V}}^{o}}^{(t,h,w)} - {{\mathbf{V}}^{b}}^{(t,h,w)} \right\|_2 > \delta \\
            0, & \text{otherwise}
        \end{cases}
    \end{equation}
where \( \delta > 0 \) is a fixed threshold ($\delta=0.09$ in our implementation). The resulting binary mask highlights pixel-level differences induced by object removal and is downsampled to match the latent resolution, yielding the ground-truth difference mask \( \mathbf{d}_t \in \{0,1\}^{f \times h/s \times w/s} \).

During the forward pass, we extract cross-attention maps $A$ across the Diffusion Transformer (DiT) blocks and reshape them into a 3D spatiotemporal map:
    \begin{equation}
        \hat{\mathbf{d}}_t = \text{Interpolate}\left(\text{Reshape}(A), \text{size} = (F, H, W)\right),
    \end{equation}

where the predicted effect map $\hat{\mathbf{d}}_t \in [0,1]^{B \times 1 \times F \times H \times W}$ is upsampled via trilinear interpolation from the coarse patch grid to the full resolution.
The module is trained under MSE loss supervision against $\mathbf{d}_t$, providing a self-localization signal that enhances sensitivity to subtle contextual updates.

During training, the model is jointly optimized:
\begin{equation}
\mathcal{L}_\text{total} =
 \mathcal{L}_{\text{diffusion}}
+ \lambda \, \mathcal{L}_\text{{EE}},
\end{equation}
where the EE term is weighted by $\lambda$.

\subsection{Three-Stage Distillation Pipeline}

\input{figure/tex/distill}

Existing video erasing diffusion models typically require around 50 denoising steps to generate high-quality video and presume that the entire input video and mask is known before generation.
To enable streaming erasing,
we propose a three-stage distillation pipeline that effectively transfers the high-fidelity erasing capabilities of a foundation model into an ultra-fast, unbidirectional streaming erasor. An overview of the framework is illustrated in Fig.~\ref{distill}.

\noindent \textbf{Stage 1: Bidirectional Teacher Training.}
To establish a powerful offline erasing foundation, we first train a bidirectional teacher model $\epsilon_{\theta}^{\text{bid}}$ using the full temporal-spatial attention mechanism over 50 denoising steps. Building upon the Wan2.1 backbone\citep{wan2025wan}, the model is optimized using $\mathcal{L}_{\text{total}}$ defined above. 
This stage endows the foundation model with versatile video erasing capabilities and side-effect awareness, serving as a high-fidelity teacher for subsequent distillation.

\noindent \textbf{Stage 2: Causal Student Initialization.}
To enable streaming video erasing, the architecture must transition from a bidirectional attention paradigm to an AR streaming formulation. 
Inspired by recent causal video generation frameworks~\citep{zhu2026causal}, we adopt a chunk-wise causal attention design $M_{\text{causal}}$ that restricts temporal tokens from attending to future video chunks, establishing a Teacher Forcing mechanism essential for streaming synthesis. 
Following the initialization protocol in Causal Forcing, we adapt the causal model for video erasing and distill it into a four-step causal student by regressing on ODE solution pairs sampled from the teacher. 
The student generator $\epsilon_{\theta}^{\text{causal}}$ is optimized via the same erasing objective $\mathcal{L}_{\text{total}}(M_{\text{causal}})$.
This stage produces a robust four-step causal student generator, serving as the initialization for subsequent distillation.

\noindent \textbf{Stage 3: DMD One-Step Distillation.}
To achieve ultra-low latency while mitigating accumulated shift errors during continuous streaming (illustrated in Fig.~\ref{appendix:stage}), we perform advanced step-distillation using Distribution Matching Distillation (DMD).
Recent studies on one-step streaming video diffusion focus on video generation, typically requiring clean past frames as input for motion plausibility. Teacher forcing (conditioning on ground truth)~\citep{lin2025autoregressive} causes error accumulation during inference, while student forcing (conditioning on predicted latents)~\citep{huang2026self} alleviates this but requires sequential unfolding, reducing efficiency.

Specifically, we initialize our 1-step DMD generator $G_{\theta}$ using the 4-step causal student parameters ($\epsilon_{\theta}^{\text{causal}}$) from Stage 2. 
The Stage~1 full-attention DiT serves as the frozen Real Score function $\epsilon_{\phi}^{\text{real}}$, while a trainable Fake Score model $\epsilon_{\psi}^{\text{fake}}$ is initialized to model the distribution of generated latents.
The generator $G_{\theta}$ receives the conditioned inputs alongside Gaussian noise, operating under the chunk-wise causal mask $M_{\text{causal}}$ at a single timestep. 
The distillation process is jointly optimized by $\mathcal{L}_{MSE}$ and the DMD gradient $\nabla_{\theta}\mathcal{L}_{DMD}$, which is computed between $\epsilon_{\phi}^{real}$ and  $\epsilon_{\psi}^{fake}$:
\begin{equation}
\nabla_{\theta}\mathcal{L}_{\text{DMD}} = \mathbb{E}_{\boldsymbol{z}_T, \boldsymbol{c}} \left[ w(t) \left( \epsilon_{\phi}^{\text{real}}(\boldsymbol{z}_t, t, \boldsymbol{c}) - \epsilon_{\psi}^{\text{fake}}(\boldsymbol{z}_t, t, \boldsymbol{c}) \right) \nabla_{\theta} G_{\theta}(\boldsymbol{z}_T, \boldsymbol{c}) \right],
\end{equation}
where $z_t$ is the intermediate latent simulated from the generated output $G_{\theta}(z_T,c)$, and $w(t)$ is a timestep-dependent weighting function. 

Crucially, this design resolves the train--infer gap. By discarding AR dependencies on previously generated latents and performing one-step generation conditioned on input video, mask and noise, the model avoids feeding its own prediction errors back into future steps.
Temporal continuity is instead maintained implicitly via the KV-cache, where clean latent representations from deeper layers propagate across streaming chunks. Consequently, $G_{\theta}$ focuses purely on localized content restoration while enabling high-fidelity, single-step video erasing in real-time.

\subsection{Long Video Streaming Generation}
\noindent\textbf{Attention sink.}\quad
Autoregressive generation with sliding-window attention, as adopted in Self Forcing~\citep{huang2026self}, is prone to quality degradation and drift during long-video extrapolation. 
Prior work in large language models has shown that initial tokens play a crucial role in stable streaming generation~\citep{xiao2023efficient}.
MotionStream~\citep{shin2026motionstreamrealtimevideogeneration} further observes a similar phenomenon in video diffusion models.
Inspired by this, we adapt the attention sink concept to our video erasing model by maintaining a local attention window while 
preserving tokens from the initial frame as a fixed anchor throughout generation.
In our chunk-wise autoregressive setup, this sinking mechanism prevents the model from drifting during long-video extrapolation, even when trained only on short sequences limited by the teacher model.

\noindent \textbf{Inference.}\quad
To maintain continuous streaming generation over long videos without quality degradation, our inference procedure strictly mirrors the training setup to achieve an exact train-test alignment.
Specifically, we maintain a KV cache containing $S$ chunks from the initial frames and a fixed-size local window of recent $W$ chunks. As new tokens are generated, this local window is rolled forward to maintain a constant size. For positional encoding, we store the final, RoPE-applied values for the static sink tokens, while tokens in the rolling window store pre-RoPE activations and receive positional indices dynamically based on their current cache location. Because this entire mechanism is simulated during training, the model seamlessly handles the discontinuity between sink and window tokens. We further analyze the effects of attention sink and window size in Sec~\ref{abla_long}.

%% file: figure/tex/method1.tex
\begin{figure*}[t]
    \centering
    \includegraphics[width=\textwidth]{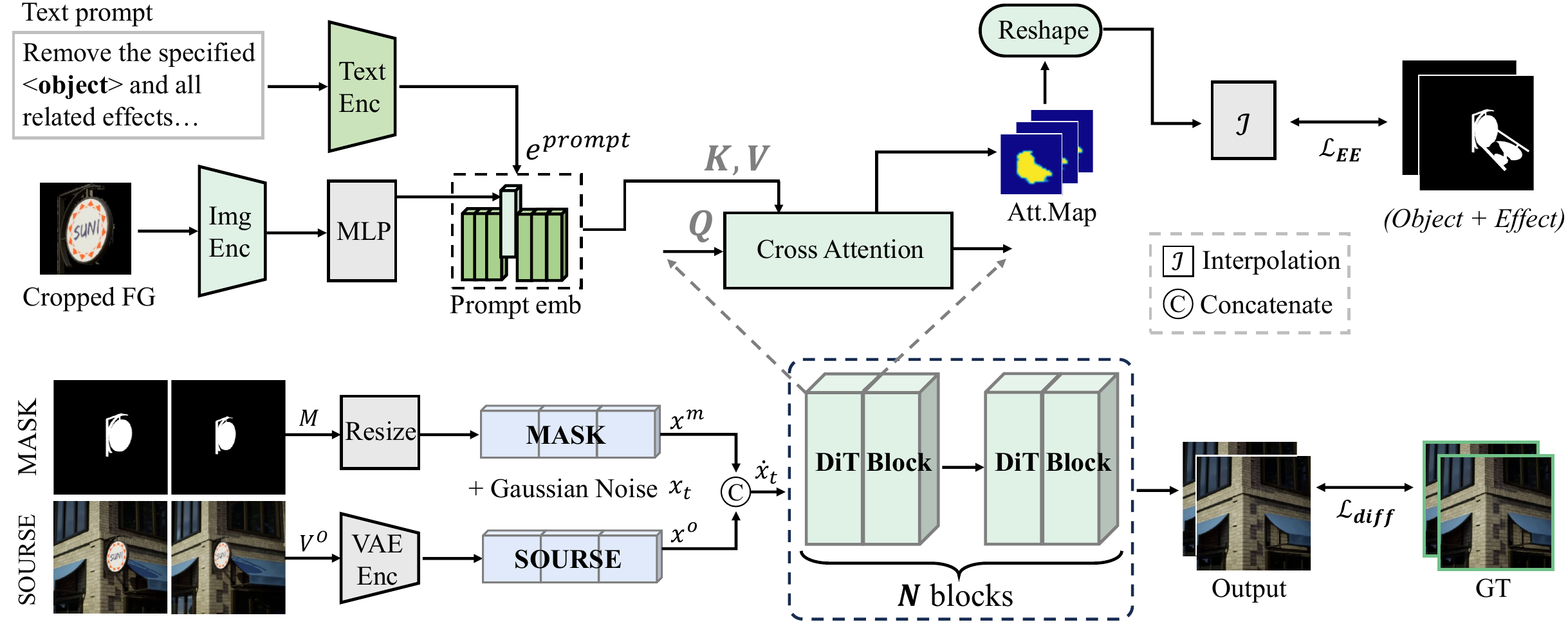}
    \vspace{-0.8cm}
    \caption{\textbf{The model architecture of \method.} We concatenate the noisy latents with the original input video and masks, consumed by a video inpainting model.
    Each DiT block performs cross-attention using the fused features $\dot{\boldsymbol{x}}_t$ as Query and $\boldsymbol{e}^{\text{prompt}}$ from Region Guidance as Key/Value, producing attention maps that highlight affected regions.
    We aggregate attention maps from all blocks, which is supervised by the effect erasing loss $\mathcal{L}_{\text{EE}}$ to focus on the affected area.
    }
    \label{method}
    \vspace{-0.5cm}
\end{figure*}

%% file: figure/tex/distill.tex
\begin{figure*}[t]
    \centering
    \includegraphics[width=\textwidth]{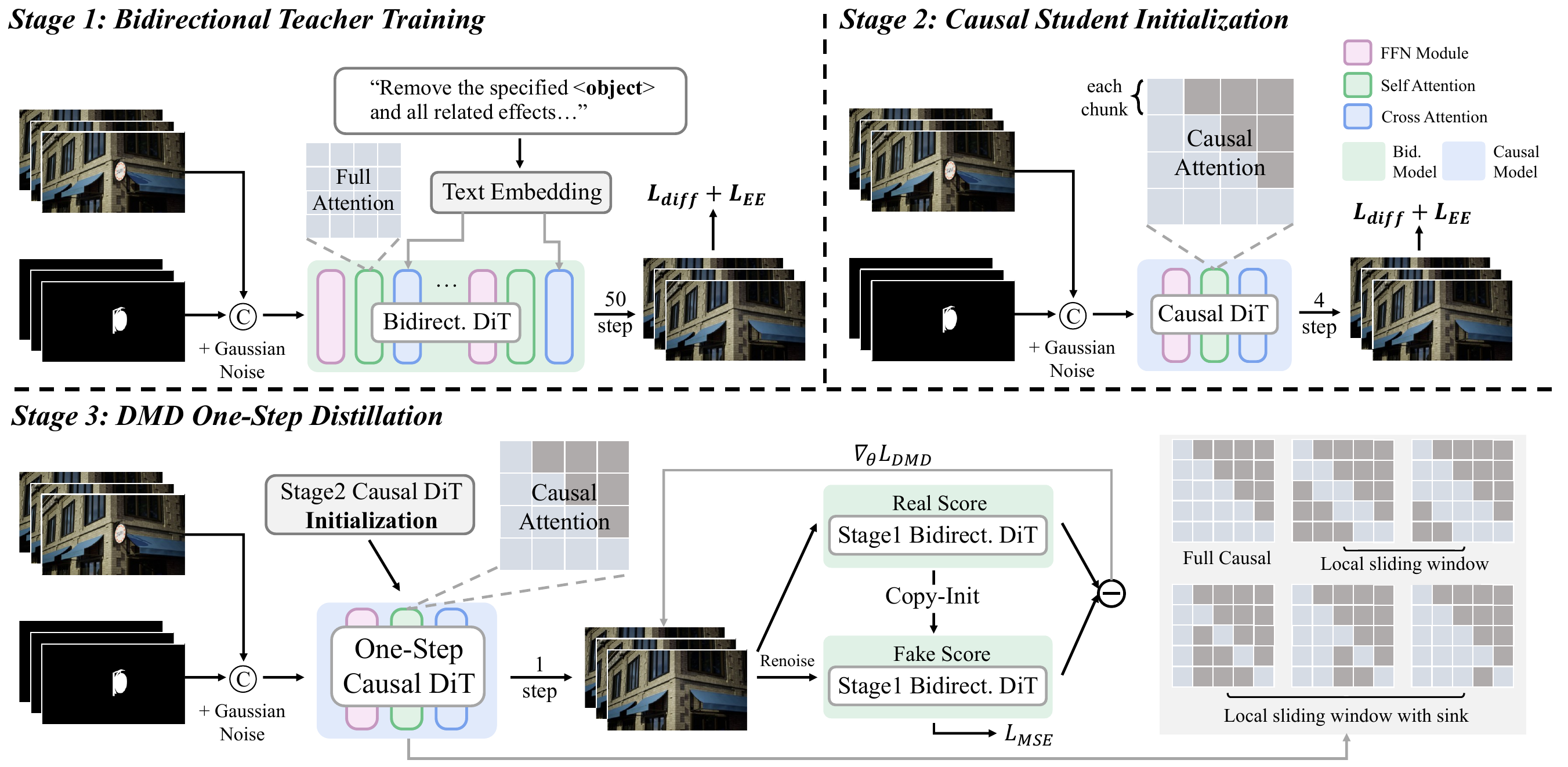}
    \vspace{-0.8cm}
    \caption{\textbf{Overview of the proposed streaming video erasing framework.} Our approach features a three-stage distillation pipeline that transfers erasing capabilities from a bidirectional teacher model to a 1-step causal student model. Each stage utilizes the backbone detailed in Fig.~\ref{method}.
    }
    \label{distill}
    \vspace{-0.6cm}
\end{figure*}

%% file: section/5_exp.tex
\section{Experiments}
\input{table/table1}

\noindent \textbf{Implementation Details.}\quad
Our method is built on the Wan 1.3B 2.1~\cite{wan2025wan} video generation model.
\label{details}
The input resolution is set to $832 \times 480$, and 81 consecutive frames are randomly sampled for training.
For training, our dataset comprises 16K high-quality video-video pairs, which are carefully filtered from the large-scale ROSE~\citep{miao2025rose} dataset.
All training is conducted on $8 \times$ A100 GPUs, while inference is performed on H200. See more details in Appendix~\ref{traning_details}

\noindent \textbf{Evaluation Data.}\quad
We evaluate \method{} against existing methods on three datasets:  
(1) \textit{ROSE-Benchmark}, a synthetic dataset with 60 paired videos;
(2) \textit{VOR-Eval}: a real-world dataset comprising 43 video pairs;
\textit{(3) \textit{LongErase}}, our newly proposed benchmark for long-video object removal. 
Existing benchmarks focus primarily on short clips due to the bidirectional attention constraints of diffusion models.
To bridge this gap, we introduce \textit{LongErase}, a challenging real-world benchmark consisting of 80 video sequences, each 30 seconds long, complete with ground-truth counterparts.

\noindent \textbf{Evaluation Metrics.}\quad
For ROSE and VOR-Eval, we adopt standard fidelity metrics, including PSNR~\citep{hore2010image}, SSIM~\citep{wang2004image}, LPIPS~\citep{zhang2018unreasonable}, and FVD~\citep{unterthiner2019fvd}.
\input{figure/tex/visual}
For LongErase, we evaluate multi-dimensional video quality using VBench~\citep{huang2023vbench}.
Furthermore, to assess quality drift in long video generation, we compute the absolute difference in imaging quality, $\delta_{Drift}^{Quality}$ , between the first and the last 5-second segments of each video. The magnitude of $\delta_{Drift}^{Quality}$ reflects the severity of error accumulation.

\subsection{Comparison with State-of-the-Art Methods.} 
We compare \method{} with several state-of-the-art image inpainting methods~\citep{zhao2025objectclear,yu2025omnipaint} applied in a per-frame manner, video inpainting methods~\citep{vace,zhou2023propainter,li2025diffueraser}, and advanced video object removal methods~\citep{zi2025minimaxremovertamingbadnoise,miao2025rose,fu2026effecterase}.
\input{table/table2}%
\noindent \textbf{Quantitative Evaluation.}\quad 
As shown in Tab.~\ref{tab1}, current image inpainting methods~\citep{zhao2025objectclear,yu2025omnipaint} process video frames independently using 2D models, and thus struggle to preserve temporal consistency across frames.
Recent video inpainting methods~\citep{zhou2023propainter,vace,li2025diffueraser} do not explicitly model object side effects, leading to unnatural removal outcomes.
Existing video object removal methods~\citep{miao2025rose,fu2026effecterase} lack spatiotemporal correlation modeling between the object and its side effects, and consequently often produce artifacts and residual traces of the removed objects.
Overall, \method{} consistently achieves strong performance on both ROSE-Benchmark and VOR-Eval while operating at 26.06 FPS, substantially improving the efficiency over existing methods. 
Beyond standard short-video evaluation, Tab.~\ref{tab:vbench_full} assesses \method{} on both 5-second and 30-second videos. \method{} remains highly competitive on short clips and maintains comparable visual quality when extended to 30 seconds with minimal degradation, demonstrating its scalability to longer sequences.

\noindent \textbf{Qualitative Evaluation.}\quad Qualitative comparisons are  presented  in Fig.~\ref{visual}.  
Video inpainting methods~\citep{vace,zi2025minimaxremovertamingbadnoise} often struggle to completely remove the target objects and their associated effects, leaving noticeable residuals and artifacts in the edited regions.
Previous object removal approaches, such as ROSE~\citep{miao2025rose} and EffectErase~\citep{fu2026effecterase}, achieve more effective object removal, but tend to suffer from flickering and blurring artifacts in highly dynamic scenarios with substantial object or camera motion.
In contrast, \method{} effectively removes target objects and their associated effects, while maintaining temporal consistency, producing clean, coherent, and high-quality results.
Visual results on long videos are show in Fig.~\ref{long_demo}.

\subsection{Ablation Studies}
\label{abla_long}

\noindent\textbf{Three-Stage Distillation Pipeline.}\quad
We conduct ablation experiments to assess the contribution of each stage in our pipeline, as summarized in Tab.~\ref{ablation}.

\noindent \textit{Stage 1: High-Fidelity Foundation.}\quad 
Stage~1 establishes a high-fidelity foundation using full temporal-spatial attention. This stage provides comprehensive erasing capabilities, but its bidirectional formulation relies on Classifier-Free Guidance (CFG) with 100 Network Function Evaluations (NFEs), incurring heavy computational overhead and precluding efficient streaming inference.

\noindent \textit{Stage 2: Causal Streaming Adaptation.}\quad
Stage~2 converts the foundation model into a streaming model through causal attention over sequential 3-frame chunks. This enables the desired streaming input-output paradigm while reducing the inference cost to 8 NFEs per chunk. 
Despite achieving streaming capability, this autoregressive step still exhibits noticeable error accumulation across chunks, primarily manifest as progressive color shifts and visual degradation.

\input{table/abla1}

\noindent \textit{Stage 3: One-Step stability.}\quad
Stage~3 resolves the remaining efficiency and stability bottlenecks. Through one-step DMD, the generator reduces inference to a single NFE while avoiding iterative latent propagation. The distilled model preserves the high-fidelity erasing behavior from Stage~1 while maintaining the streaming capability introduced in Stage~2. As a result, it enables ultra-low-latency video erasing with substantially improved stability and no observable error accumulation.

\noindent\textbf{Model Components.}\quad
We further investigate the key design choices in our method.

\noindent \textit{Effectiveness of Effect Erasing Loss.}\quad
The proposed EE loss encourages model focus on side-effect regions, strengthening its attention to the affected areas.
As shown in the Tab.~\ref{tab:abla2}, adding the EE loss consistently improves the baseline across all metrics, with PSNR increasing from 30.104 to 32.571.

\noindent \textit{Effectiveness of Multimodal Region Guidance.}\quad
The MMRG module captures spatiotemporal correlations between objects and their side effects, enabling the model to localize and perceive affected regions.
As shown in Tab.~\ref{tab:abla2}, MMRG enables the model to produce higher-quality erasure results, with SSIM improving significantly from 0.818 to 0.872.
\input{figure/tex/long_abla}

\subsection{Long-Video Streaming Analysis}

We investigate three key design choices governing streaming quality and interactivity: the latent chunk size, the attention sink size, and the local context window size.

\noindent\textbf{Impact of Chunk Size.}\quad
The chunk size, which defines how many video latents are processed in parallel, presents a critical trade-off between quality, latency, and throughput. 
As shown in Fig.~\ref{appendix:abla}, small chunks yield lower throughput, while large chunks ($>$3) cause prohibitive latency for real-time interaction.
Furthermore, Tab.~\ref{tab:long} shows that a chunk size of 1 severely degrades quality.
We thus select 3 as our default chunk size (see Appendix~\ref{sec:appendix_abl} for details).
\input{table/abla2}%
\noindent\textbf{Attention Sink and Context Window.}\quad
To evaluate attention sinks and window sizes, we train a single model with randomly sampled configurations and test long-video extrapolation on LongErase (average 486 frames).
Surprisingly, we find that the minimal configuration (a single-chunk sink with a single-chunk window) achieves the best performance. 
Fig.~\ref{long_abla}(a) shows that additional sinks offer marginal gains while increasing latency, and expanded windows degrade performance. 
We hypothesize that restricting context to immediate predecessors mitigates error accumulation and reduces drift for long-video generation.

We additionally show reconstruction accuracy and throughput as well as their time evolution in Tab.~\ref{tab:long} and Fig.~\ref{long_abla}(b,c) respectively.
We denote our default configuration (\texttt{chunk-3, sink-1, window-1}) as \texttt{c3s1w1}.
In contrast, Self Forcing uses a sliding window approach (\texttt{c3s0w6}) that attends to up to 6 previous chunks without sink tokens and employs unbounded RoPE positions scaling with frame count.
Although removing sink tokens (\texttt{c3s0w1}) yields marginal speed improvements, this  comes at the cost of degraded long-term generation stability.

\input{figure/tex/long_demo}

We also develop an interactive demo that supporting real-time user control during streaming generation. Fig.~\ref{long_demo} shows an example.

%% file: table/table1.tex
\begin{table*}[t]
\centering
\footnotesize
\caption{\textbf{Quantitative results on ROSE and VOR-Eval.} 
The best and second-best results are highlighted in bold and underlined, respectively.}
\label{tab1}
\renewcommand{\arraystretch}{1.06}
\setlength{\tabcolsep}{4.5pt}
\begin{tabular}{lccccccccc}
\toprule
\multirow{2}{*}{Method} & \multirow{2}{*}{FPS} & \multicolumn{4}{c}{ROSE-Benchmark} & \multicolumn{4}{c}{VOR-Eval} \\
\cmidrule(lr){3-6} \cmidrule(lr){7-10}
& & PSNR↑ & SSIM↑ & LPIPS↓ & FVD↓ & PSNR↑ & SSIM↑ & LPIPS↓ & FVD↓ \\
\midrule
ObjectClear      & * & 29.535 & 0.920 & 0.076 & 742.829 & 22.403 & 0.787 & \textbf{0.190} & 1391.858 \\ 
OmniPaint        & * & 27.569 & 0.910 & 0.085 & 809.645 & 21.511 & 0.781 & 0.201 & 1439.867 \\ 
\midrule
Propainter       & 3.92 & 27.200 & 0.915 & 0.095 & 171.020 & 21.975 & 0.800 & 0.225 & \ \ 589.012 \\
DiffuEraser      & 1.93 & 26.502 & 0.898 & 0.128 & 167.483 & 21.946 & 0.802 & 0.214 & \ \ 559.497 \\
VACE             & 0.57 & 20.805 & 0.694 & 0.174 & 254.117 & 17.677 & 0.591 & 0.294 & \ \ 806.476 \\ 
\midrule
MiniMax-Remover  & \underline{5.36} & 26.770 & 0.905 & 0.099 & 137.840 & 21.963 & 0.802 & 0.217 & \ \ 539.427 \\
ROSE             & 1.05 & 30.759 & 0.915 & 0.079 & 81.218 & 21.752 & 0.781 & 0.216 & \ \ 403.315 \\
EffectErase      & 1.51 & \underline{31.542} & \underline{0.921} & \underline{0.068} & \underline{59.723} & \textbf{22.792} & \underline{0.803} & \underline{0.196} & \ \ \textbf{335.497} \\
\rowcolor[gray]{0.95}\textbf{\method} & \textbf{26.06} & \textbf{32.791} & \textbf{0.928} & \textbf{0.063} & \textbf{56.294} & \underline{22.494} & \textbf{0.805} & 0.206 & \ \ \underline{344.846} \\
\bottomrule
\end{tabular}
\vspace{-0.6cm}
\end{table*}

%% file: figure/tex/visual.tex
\begin{figure*}[t]
    \centering
    \includegraphics[width=\textwidth]{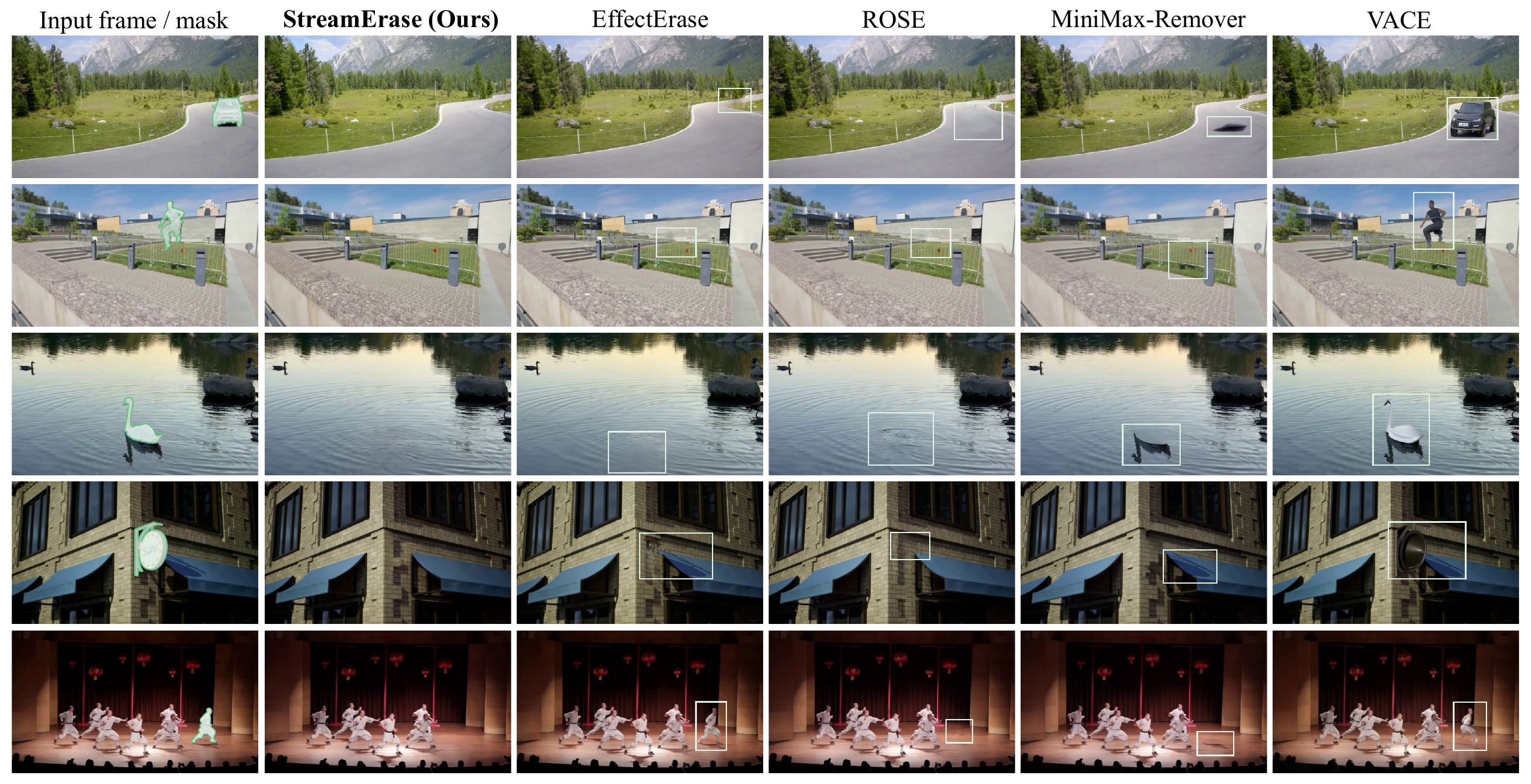}
    \vspace{-0.9cm}
    \caption{\textbf{Qualitative comparison between \method{} and existing approaches on real-world samples.} Our model demonstrates superior ability and effectively handles complex object-environment interactions, including shadows, occlusions, and illumination changes.}
    \label{visual}
    \vspace{-0.7cm}
\end{figure*}

%% file: table/table2.tex
\begin{wraptable}{r}{0.55\columnwidth}
    \vspace{-2em}
    \centering
    \caption{\textbf{Quantitative evaluation on LongErase.} Comparison on 5-second clips and long-term performance on 30-second videos. The best score is bolded.}
    \label{tab:vbench_full}
    \setlength{\tabcolsep}{2pt} 
    \small
    \resizebox{\linewidth}{!}{%
    \begin{tabular}{lcccccc}
    \toprule
    \multirow{2}{*}{Method} & \multicolumn{5}{c}{Evaluation Scores $\uparrow$} & \multirow{2}{*}{\makecell{$\mathbf{\delta_{Drift}^{Quality} \downarrow}$}} \\
    \cmidrule(lr){2-6}
    & \makecell{Motion\\ Smoothness} 
    & \makecell{Background\\ Consistency} 
    & \makecell{Temporal\\ Flickering} 
    & \makecell{Subject\\ Consistency} 
    & \makecell{Imaging\\ Quality} & \\
    \midrule
    \multicolumn{7}{c}{\textit{Short-Video Comparison (5s)}} \\
    \midrule
    DiffuEraser               & 0.968 & 0.917 & \textbf{0.939} & 0.947 & \textbf{0.614} & - \\
    ProPainter                & 0.972 & 0.925 & 0.935 & 0.951 & 0.602 & - \\
    ROSE                      & 0.971 & 0.933 & 0.922 & 0.955 & 0.602 & - \\
    EffectErase               & \textbf{0.987} & 0.937 & 0.937 & 0.956 & 0.604 & - \\
    \rowcolor[gray]{0.95} \textbf{\method}     & 0.986 & \textbf{0.938} & \textbf{0.939} & \textbf{0.960} & 0.610 & - \\
    \midrule
    \multicolumn{7}{c}{\textit{Long-Video Evaluation (30s)}} \\
    \midrule
    \rowcolor[gray]{0.95} \textbf{\method}     & 0.987     & 0.936     & 0.940     & 0.958     & 0.608     & 0.05 \\
    \bottomrule
    \end{tabular}%
    }
    \vspace{-0.6em}
\end{wraptable}

%% file: table/abla1.tex
\begin{table*}[t]
\centering
\begin{minipage}[t]{0.48\textwidth}
  \centering
  \caption{\textbf{Ablation of our three-stage distillation pipeline.} Comparison of generation configurations and efficiency across the different stages. All metrics are evaluated on ROSE.}
  \label{ablation}
  \resizebox{\linewidth}{!}{%
    \begin{tabular}{lccc}
      \toprule
      & \textbf{Stage 1 (Foundation)} & \textbf{Stage 2 (Streaming)} & \textbf{Stage 3 (DMD)} \\
      \midrule
      Is streaming? & $\times$ & $\checkmark$ & $\checkmark$ \\
      NFEs & 100 & 8 & 1 \\
      With CFG? &  $\checkmark$ & $\checkmark$ & $\times$  \\
      Latency & 130.16 & 1.35 & 0.87 \\
      Chunk size & Full sequence & 3 frames & 3 frames \\
      \midrule
      PSNR & 32.487 & 30.183 & 32.791 \\
      SSIM & 0.925 & 0.903 & 0.928 \\
      LPIPS & 0.065 & 0.084 & 0.063 \\
      \bottomrule
    \end{tabular}%
  }
\end{minipage}%
\hfill 
\begin{minipage}[t]{0.48\textwidth}
  \centering
  \caption{\textbf{Ablation study on ROSE.} Based on ROSE data, the removal performance improves progressively with the addition of the effect erasing loss ($\mathcal{L}_\text{{EE}}$), Multimodal Region Guidance (MMRG).}
  \label{tab:abla2}
  \renewcommand{\arraystretch}{1.2}
  \resizebox{\linewidth}{!}{%
    \begin{tabular}{lcc|cccc}
      \toprule
      Exp. & $\mathcal{L}_\text{{EE}}$ & MMRG & PSNR$\uparrow$ & SSIM$\uparrow$ & LPIPS$\downarrow$ & FVD$\downarrow$ \\
      \midrule
      (a) & & & 30.104 & 0.818 & 0.089 & 108.664 \\
      (b) & $\checkmark$ & & 32.571 & 0.894 & 0.073 & 72.544 \\
      (c) & & $\checkmark$ & 31.613 & 0.872 & 0.084 & 84.147 \\
      \rowcolor[gray]{0.95}(d) & $\checkmark$ & $\checkmark$ & \textbf{32.791} & \textbf{0.928} & \textbf{0.063} & \textbf{56.294} \\
      \bottomrule
    \end{tabular}%
  }
\end{minipage}
\vspace{-0.6cm}
\end{table*}

%% file: figure/tex/long_abla.tex
\begin{figure*}[t]
    \centering
    \includegraphics[width=\textwidth]{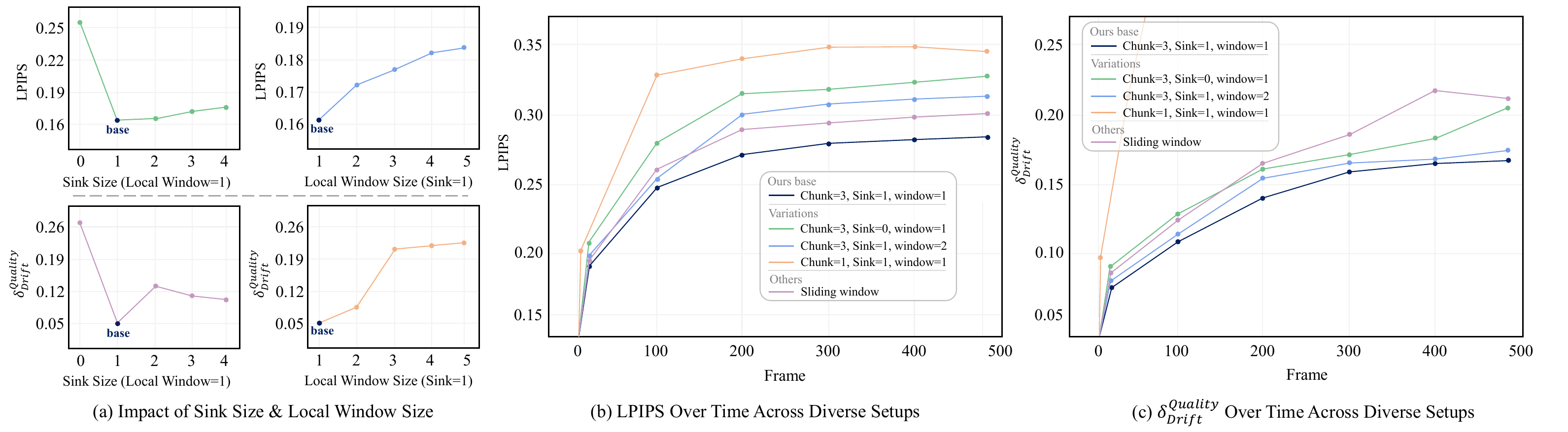}
    \vspace{-0.8cm}
    \caption{
    \textbf{Impact of Sparse Attention Patterns.} Using longer clips (average 486 frames) from the LongErase, we ablate attention sink size and local window size in extrapolation scenarios. Having at least a single sink chunk is crucial, but more provides marginal benefit, while larger window sizes degrade performance as attending to long-past history allows errors to accumulate in context tokens.
    }
    \label{long_abla}
    \vspace{-0.6cm}
\end{figure*}

%% file: table/abla2.tex
\begin{wraptable}{r}{0.48\columnwidth}
  \vspace{-2em}
  \caption{\textbf{Long-video streaming design analysis on LongErase.}
\texttt{c3s1w1} achieves the best visual quality and long-term stability.}
  \label{tab:long}
  \renewcommand{\arraystretch}{1.15}
  \resizebox{\linewidth}{!}{%
    \begin{tabular}{lcccc}
      \toprule
      \multirow{2}{*}{Config} & \multicolumn{4}{c}{LongErase} \\
      \cmidrule(lr){2-5}
      & LPIPS & $\mathbf{\delta_{Drift}^{Quality}}$ & Latency (s) & FPS \\
      \midrule
      Ours base (\texttt{c3s1w1}) & \textbf{0.163} & \textbf{0.05} & 0.87 $\pm$ 0.04 & \underline{26.06 $\pm$ 1.62} \\
      + Remove sink (\texttt{c3s0w1}) & 0.254 & 0.27 & \underline{0.85 $\pm$ 0.01} & \textbf{27.28 $\pm$ 1.70}  \\
      + Chunk Size 1 (\texttt{c1s1w1}) & 0.319 & 0.72 & \textbf{0.42 $\pm$ 0.03} & 20.70 $\pm$ 2.97 \\
      Sliding window & \underline{0.181} & \underline{0.08} & 0.91 $\pm$ 0.15 & 22.69 $\pm$ 3.29 \\
      \bottomrule
    \end{tabular}%
  }
  \vspace{-0.6em}
\end{wraptable}

%% file: figure/tex/long_demo.tex
\begin{figure*}[ht]
    \vspace{-0.4cm}
    \centering
    \includegraphics[width=\textwidth]{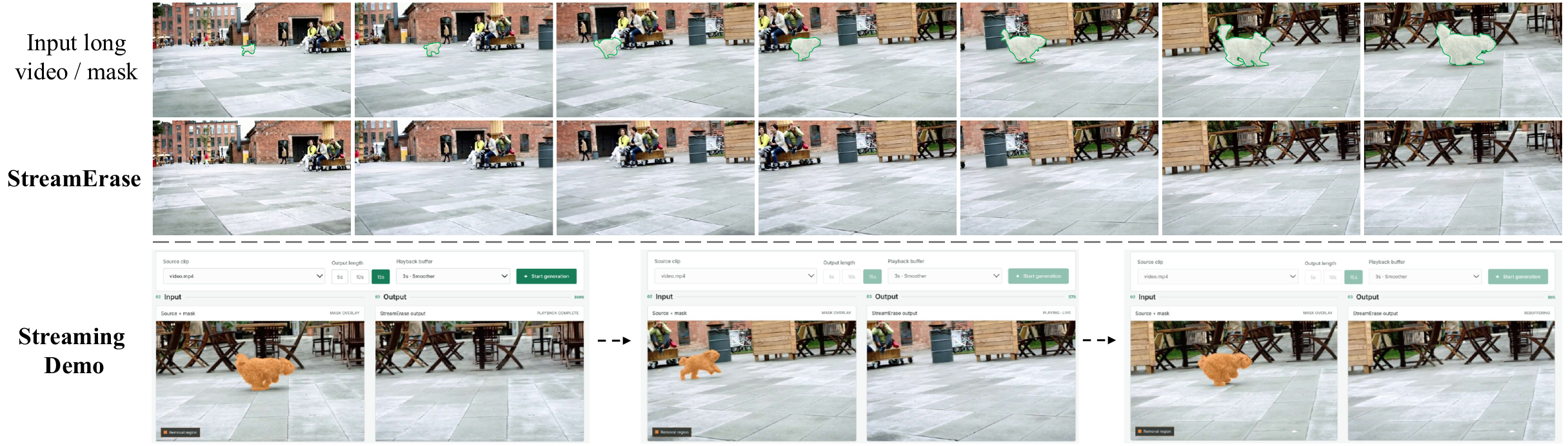}
    \vspace{-0.7cm}
    \caption{
    \textbf{Visualization of long-video eraseing and Streaming Demo.} Our method supports flexible length video erasing while preserving high visual fidelity.
    }
    \label{long_demo}
    \vspace{-0.3cm}
\end{figure*}

%% file: section/6_conclusion.tex
\section{conclusion}
In this paper, we presented \method, a novel streaming video erasing framework that successfully adapts powerful
bidirectional diffusion priors to an efficient, unidirectional
autoregressive paradigm. To overcome the attention distribution shift inherent in causal execution, we introduced a progressive three-stage distillation pipeline with attention sink and rolling KV caches. 
\method{} achieves strong performance across diverse real-world video erasing scenarios, while being significantly faster than prior methods. Limitations and future directions are discussed in Appendix~\ref{limitation}.

%% file: section/9_ethics.tex
\section*{AI Use Statement}

In this work, we used generative AI tools to assist with translation during the drafting of the paper.
We have not used generative AI tools for generating synthetic data sets, developing theoretical models or conceptual frameworks, formulating mathematical claims, providing critical ingredients for proving mathematical claims, assisting in the writing of proofs, proposing or refining hypotheses, designing or providing feedback on research methodology or experiments, implementing methods, cleaning or reformatting datasets, supporting qualitative or thematic data analysis, and interpreting results, and these tasks are not applicable to this work.
Additionally, we used generative AI tools to edit the paper to improve readability.
We have reviewed all AI-assisted work: all AI-assisted text was checked and revised by the authors for accuracy and clarity.
We take responsibility for the final content of this work, including text, claims, or artifacts produced with the aid of generative AI.

\section*{Reproducibility}
Our models are built upon publicly available Wan model variants and datasets (ROSE). Training was conducted in a well-controlled environment, with all training details, hyperparameters, and implementation specifics provided in Section~\ref{details} and the Appendix~\ref{traning_details} to ensure reproducibility.

%% file: section/8_appendix.tex
\clearpage

\section{More training details}
\label{traning_details}

\noindent \textbf{Training Setup.} The three-stage distillation pipeline is trained on 8 NVIDIA A100 GPUs. We employ the AdamW optimizer across all stages. 
\begin{itemize}
    \item \textbf{Stage 1 (Bidirectional Teacher Training):} The bidirectional foundation model $\epsilon_{\theta}^{bid}$ is trained for 16K steps with a learning rate of $2 \times 10^{-5}$ and a global batch size of 8. We utilize a standard noise scheduler with continuous timesteps $t \in [0, 1000]$.
    \item \textbf{Stage 2 (Causal Student Initialization):} The causal DiT $\epsilon_{\theta}^{causal}$ is trained for an additional 16K steps  with a learning rate of $2 \times 10^{-5}$. The temporal chunk size is set to 3 latent frames, ensuring that tokens can only attend to the current and strictly preceding chunks.
    \item \textbf{Stage 3 (DMD):} The 1-step generator $G_{\theta}$ is initialized directly from the Stage 2 weights. For the 1-step generation, the sampling timesteps are specifically set to $[1000]$. Both the Real Score model and the Fake Score model are initialized from the teacher model weights obtained in Stage 1. We apply a reduced learning rate of $2 \times 10^{-6}$ for 1K steps, utilizing the timestep-dependent weighting function $w(t)$ as proposed in standard DMD formulations.

\end{itemize}

\section{Additional Ablation Experiments}
\label{sec:appendix_abl}

\input{figure/tex/appendix1}
\noindent\textbf{Impact of Chunk Size and Sampling Steps.} In the main paper, we primarily focused on evaluating the efficiency of diverse sparse attention patterns with attention sink size and attention window size. As shown in Table~\ref{tab:long} and Figure~\ref{long_abla}, reducing the chunk size increases the number of autoregressive rollouts required to generate the same length of video while reducing the amount of bidirectional attention, leading to worse performance. Since chunk sizes beyond 3 induce high latency, which makes them less useful for our interactive use case, we report only speed metrics for these configurations. Figure~\ref{appendix:abla} (a) visualizes the latency and throughput given different chunk sizes and sampling steps. Note that since all benchmarks are conducted with a sink size and a window size of 1, the model with chunk size 7 has lower FPS than the one with chunk size 3 because its attention sequence length is longer.

The number of sampling steps per chunk also affects this balance. We train our student model for 4-step generation, which we find to be optimal for our setup; increasing beyond 4 steps yields only marginal quality gains, while reducing to 2 steps causes a noticeable drop in quality, as shown in Figure~\ref{appendix:abla}~(b). Although trained for 4-step inference, the DMD framework allows for flexible sampling configurations at test time. Based on these findings, we use a chunk size of 3 with 1 sampling step as our primary configuration to achieve a strong balance between interactivity and high-quality video generation.

\section{Details of the training modules}
\paragraph{MLP Details.}
The MLP maps the object-image features extracted by the image encoder into the latent space required by our model. It is composed of two sequential MLP blocks: the first transforms the input embedding dimension to the output dimension, and the second further refines the representation with a residual MLP. 
Each block applies LayerNorm~\citep{ba2016layer}, a linear projection, a GELU activation~\citep{hendrycks2016gaussian}, and a second linear projection, while the second block includes a residual connection. 
A final LayerNorm is applied to stabilize the projected token.

\section{Detailed Evaluation on ROSE-Bench}

\input{table/appendix/table1}

To provide a comprehensive breakdown of our proposed model, we present detailed per-category quantitative evaluations on the ROSE benchmark in Tab.~\ref{exp_rose}. Specifically, we analyze the performance evolution across our three progressive stages:
\begin{itemize}

\item \textbf{Stage 1 (High-Fidelity Initialization)}: Establishes a strong structural foundation and delivers strong reconstruction quality across general lighting scenarios. It achieves particularly strong performance in the Common and Shadow subsets.

\item \textbf{Stage 2 (Streaming Adaptation)}: Serves as a intermediate phase to enable continuous processing capability. While it introduces a minor trade-off in reconstruction fidelity, it preserves consistent structural stability across all sub-categories.

\item \textbf{Stage 3 (DMD)}: Reclaims and further improves overall visual quality, consistently achieving the highest average metrics across the benchmark. It demonstrates remarkable gains in complex optical categories such as Light Source, Reflection, Mirror, and Translucent, significantly outperforming the preceding stages.

\end{itemize}

\section{More Qualitative Comparisons and Additional Visual Results}

\input{figure/tex/appendix2}
\input{figure/tex/appendix3}

In this section, we provide additional visual results to comprehensively demonstrate the performance of \method. 
Specifically, Figure~\ref{appendix:visual} presents extended qualitative comparisons between our approach and other baseline models, highlighting our superior erasing quality and temporal consistency. 
Furthermore, Figure~\ref{appendix:stage} presents a qualitative visual comparison of video erasing performance across the three training stages. 
Figure~\ref{appendix:long} presents a qualitative comparison of long-video erasing performance.

While Stage~1 achieves superior frame-level synthesis owing to its non-causal bidirectional full attention, it lacks the temporal causality required for real-time streaming applications. 
In Stage~2, the transition to a chunk-wise causal attention architecture enables causal streaming generation. However, because inference relies on autoregressively conditioning on generated latents under a teacher-forcing distillation formulation, small discrepancy errors inevitably aggregate over successive chunks. 
As illustrated in the third column of Figure~\ref{appendix:stage}, this leads to severe error accumulation and temporal drift over time, manifested as color shift, boundary blurriness, and dynamic structure degradation (e.g., distorted backgrounds and blurred moving subjects). 

Crucially, Stage~3 resolves this error propagation bottleneck. 
By employing Distribution Matching Distillation (DMD) with the Stage~1 full-attention DiT acting as the real-score teacher, the 1-step generator $G_{\theta}$ aligns the generated latent distribution directly with the true distribution at a single timestep. 
This global distribution alignment effectively mitigates the severe drift observed in Stage~2. 
As shown in the final column, Stage~3 eliminates error accumulation, delivering sharp, visually coherent, and temporally continuous erasing results that over the quality of the Stage~1 foundation model while benefiting from ultra-low latency streaming inference.

\input{figure/tex/appendix4}

\section{Limitation and Future Work}

\label{limitation}

While \method{} enables real-time streaming video object erasing over arbitrarily long horizons, we identify several limitations. First, the fixed-context causal generation mechanism, together with attention sinks, can become less effective when the video undergoes substantial scene changes over long horizons. The model tends to preserve information from earlier contexts, which is beneficial for maintaining temporal consistency but may limit its ability to fully adapt to newly emerging scenes. Future work could explore more adaptive context management and attention mechanisms that dynamically balance historical consistency with the ability to respond to major scene transitions.

Second, the performance of \method{} is still constrained by the availability and diversity of training data for video object erasing. Existing video erasing datasets are relatively limited in scale, duration, and coverage of real-world scenarios, making it challenging to fully capture the diverse object appearances, camera motions, occlusions, and scene changes encountered in practical applications. Future work could focus on constructing large-scale, diverse, and realistic video erasing datasets with longer videos and more challenging real-world scenarios, which could further improve the robustness and generalization of streaming erasing models.

Lastly, \method{} currently relies on explicit mask inputs to specify the objects to be removed. While masks provide precise spatial control, requiring users to manually provide them introduces additional interaction cost and limits the flexibility of the system. A promising direction is to develop video erasing models that do not require explicit mask inputs, allowing users to specify unwanted objects through natural-language instructions or simple interactions such as clicking on the target object. Such a design could further reduce the burden of user input and enable a more natural and accessible interactive video erasing experience.

%% file: figure/tex/appendix1.tex
\begin{figure*}[ht]
    \centering
    \includegraphics[width=\textwidth]{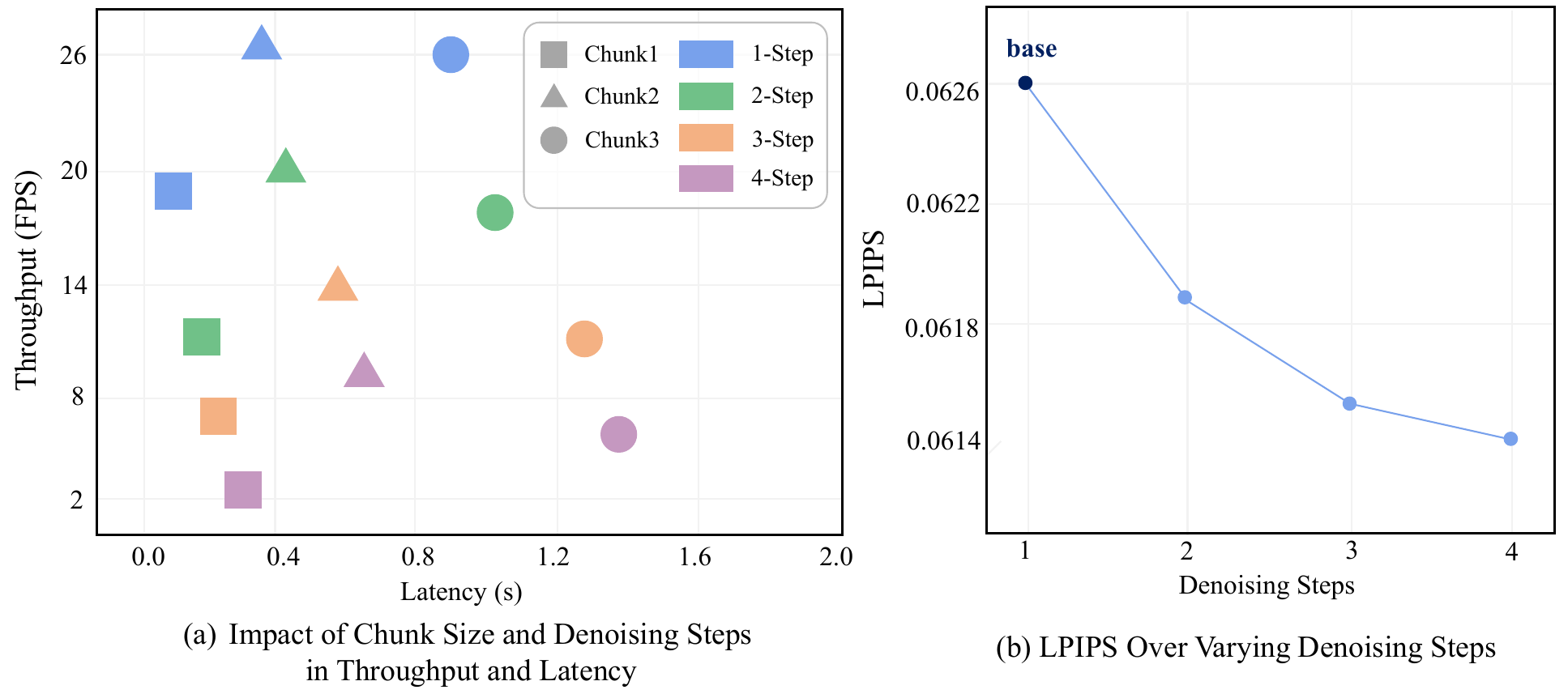}
    \caption{\textbf{Speed and Quality Tradeoffs with Chunk Sizes and Sampling Steps.} We visualize the latency-throughput relationship across varying chunk sizes and sampling steps (left), and video quality (LPIPS) across different sampling steps, using our default setup of \texttt{c3s1w1} (right).}
    \label{appendix:abla}
\end{figure*}

%% file: table/appendix/table1.tex
\begin{table}[ht]
    \centering
    \footnotesize
    \caption{Detailed quantitative comparison across subcategories on the ROSE benchmark (PSNR$\uparrow$ / SSIM$\uparrow$ / LPIPS$\downarrow$). We evaluate the progressive performance across the three stages of our method.}
    \begin{adjustbox}{max width=\textwidth}
    \begin{tabular}{ll|cccc}
    \toprule
    \textbf{Category} & \textbf{Metric} & \textbf{Stage1} & \textbf{Stage2} & \textbf{Stage3} & \textbf{ROSE} \\
    \midrule
    \multirow{3}{*}{Common} 
        & PSNR   & 36.3627 & 34.1948 & 36.2912 & \textbf{36.4248} \\
        & SSIM   & \textbf{0.9558} & 0.9359 & 0.9505 & 0.9499 \\
        & LPIPS  & \textbf{0.0332} & 0.0452 & 0.0340 & 0.0427 \\
    \midrule
    \multirow{3}{*}{Shadow}
        & PSNR   & \textbf{34.0945} & 29.8925 & 32.9684 & 33.6212 \\
        & SSIM   & \textbf{0.9253} & 0.8787 & 0.9155 & 0.9205 \\
        & LPIPS  & \textbf{0.0617} & 0.1045 & 0.0699 & 0.0644 \\
    \midrule
    \multirow{3}{*}{Light Source}
        & PSNR   & 33.2547 & 31.8069 & \textbf{34.4436} & 29.5112 \\
        & SSIM   & 0.9478 & 0.9318 & \textbf{0.9509} & 0.9159 \\
        & LPIPS  & 0.0540 & 0.0614 & \textbf{0.0450} & 0.0882 \\
    \midrule
    \multirow{3}{*}{Reflection}
        & PSNR   & 28.3543 & 26.0970 & \textbf{28.8015} & 27.8023 \\
        & SSIM   & 0.8497 & 0.8147 & \textbf{0.8565} & 0.8403 \\
        & LPIPS  & \textbf{0.1030} & 0.1405 & 0.1076 & 0.1083 \\
    \midrule
    \multirow{3}{*}{Mirror}
        & PSNR   & 31.9706 & 29.6710 & \textbf{32.3864} & 27.6502 \\
        & SSIM   & 0.9558 & 0.9392 & \textbf{0.9563} & 0.9351 \\
        & LPIPS  & 0.0608 & 0.0721 & \textbf{0.0591} & 0.0955 \\
    \midrule
    \multirow{3}{*}{Translucent}
        & PSNR   & 31.6287 & 30.8982 & \textbf{32.9972} & 30.7018 \\
        & SSIM   & 0.9564 & 0.9420 & \textbf{0.9577} & 0.9458 \\
        & LPIPS  & 0.0481 & 0.0558 & \textbf{0.0450} & 0.0620 \\
    \midrule
    \midrule
    \multirow{3}{*}{\textbf{Mean}}
        & PSNR   & 32.4865 & 30.1828 & \textbf{32.7905} & 30.7593 \\
        & SSIM   & 0.9254 & 0.9030 & \textbf{0.9284} & 0.9150 \\
        & LPIPS  & 0.0653 & 0.0836 & \textbf{0.0626} & 0.0789 \\
    \bottomrule
    \end{tabular}
    \end{adjustbox}
    \label{exp_rose}
\end{table}

%% file: figure/tex/appendix2.tex
\begin{figure*}[ht]
    \centering
    \includegraphics[width=\textwidth]{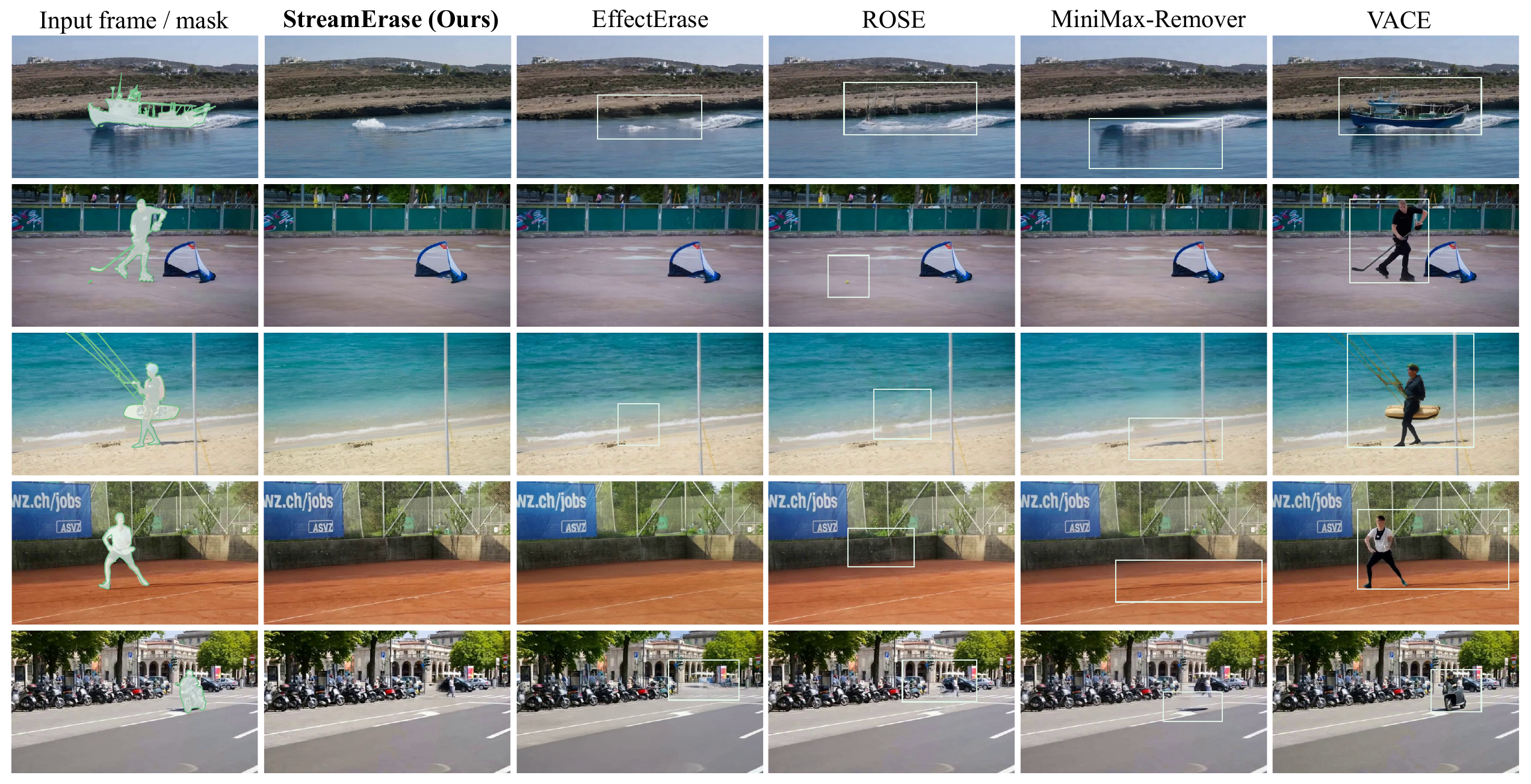}
    \caption{\textbf{More qualitative comparison between \method{} and existing approaches on real-world samples.} Our model demonstrates superior ability and effectively handles complex object-environment interactions, including shadows, occlusions, and illumination changes.}
    \label{appendix:visual}
\end{figure*}

%% file: figure/tex/appendix3.tex
\begin{figure*}[ht]
    \centering
    \includegraphics[width=\textwidth]{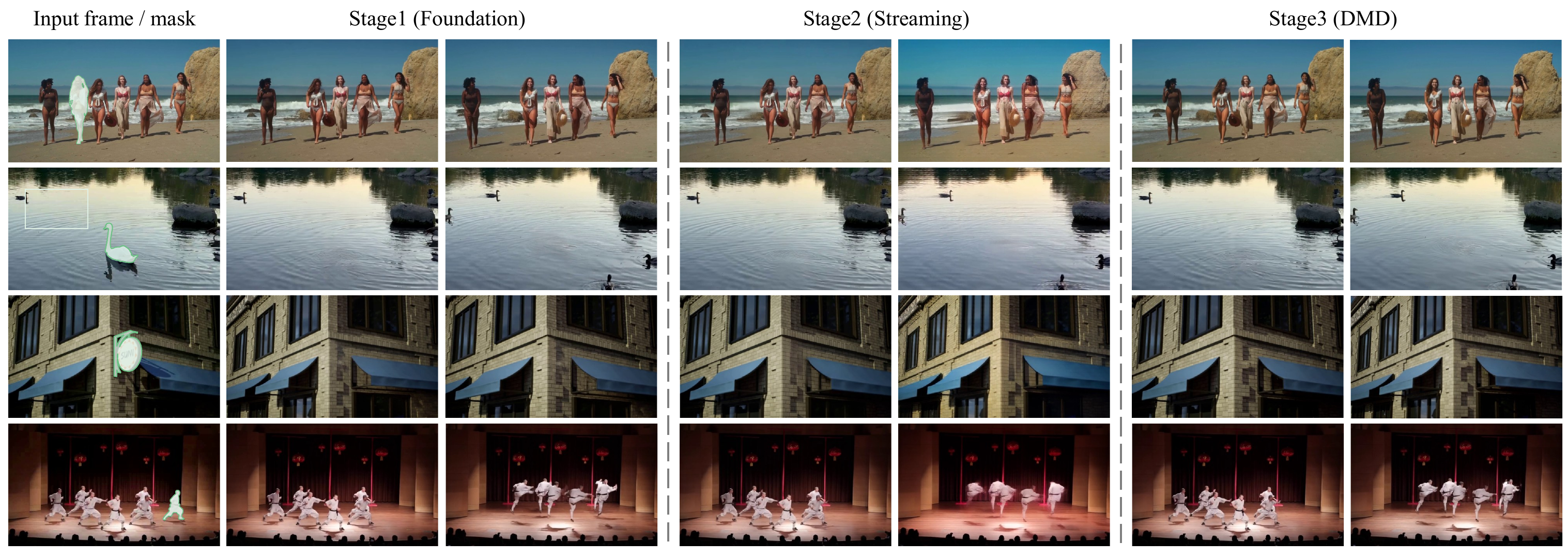}
    \caption{Qualitative comparison across training stages.}
    \label{appendix:stage}
\end{figure*}

%% file: figure/tex/appendix4.tex
\begin{figure*}[t]
    \centering
    \includegraphics[width=\textwidth]{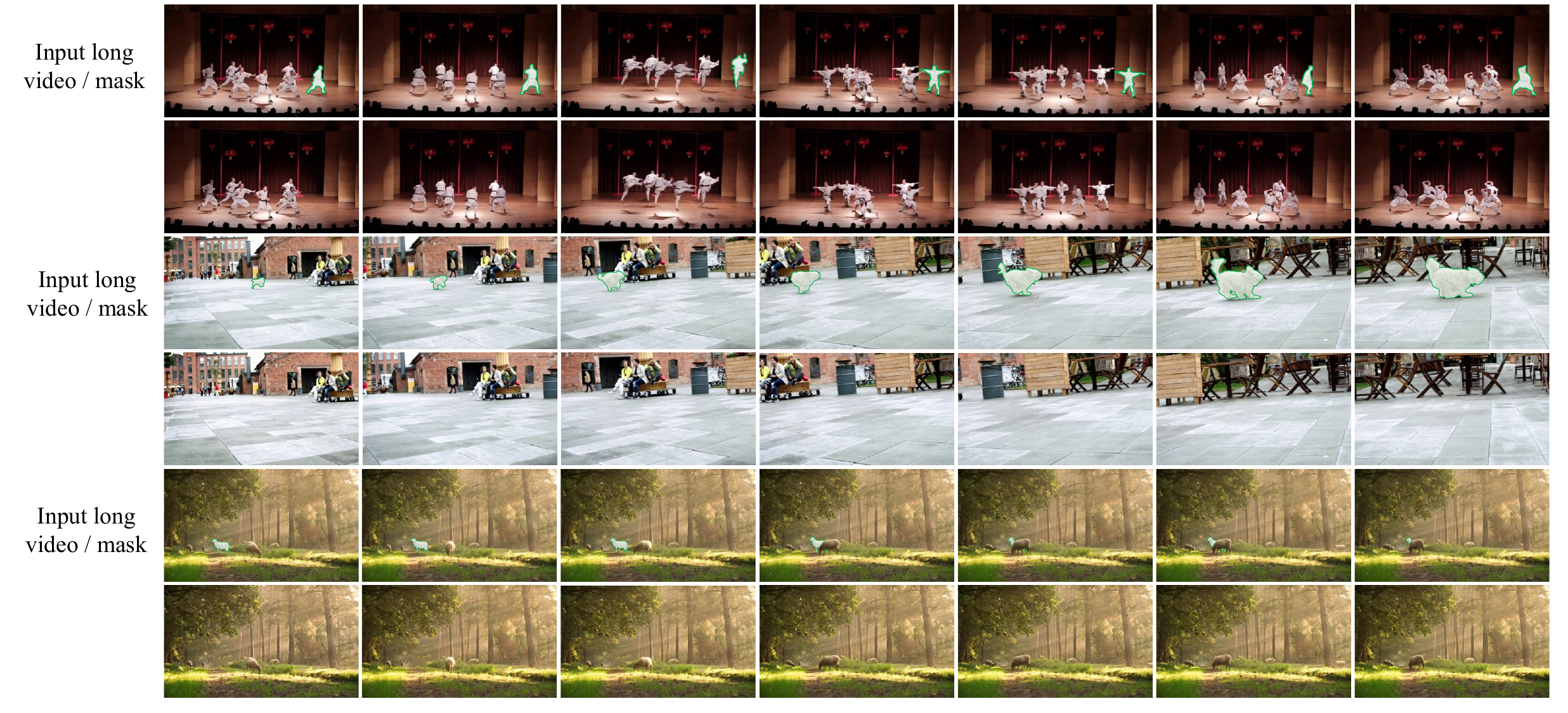}
    \caption{Qualitative Comparison of Long-Video Erasing.}
    \label{appendix:long}
\end{figure*}

%% file: main.bib
@String(CVPR= {IEEE Conf. Comput. Vis. Pattern Recog.})

@String(ICCV= {Int. Conf. Comput. Vis.})

@String(ECCV= {Eur. Conf. Comput. Vis.})

@String(NeurIPS= {Adv. Neural Inform. Process. Syst.})

@String(ICPR = {Int. Conf. Pattern Recog.})

@String(TIP  = {IEEE Trans. Image Process.})

@String(ICLR = {Int. Conf. Learn. Represent.})

@String(AAAI = {AAAI})

@String(CVPR  = {CVPR})

@String(ICCV  = {ICCV})

@String(ECCV  = {ECCV})

@String(NeurIPS  = {NeurIPS})

@String(ICPR  = {ICPR})

@String(TIP   = {IEEE TIP})

@String(ICLR  = {ICLR})

@String(ICML = {ICML})

@String(JMLR = {JMLR})

@article{li2025diffueraser,
  title={Diffueraser: A diffusion model for video inpainting},
  author={Li, Xiaowen and Xue, Haolan and Ren, Peiran and Bo, Liefeng},
  journal={arXiv preprint arXiv:2501.10018},
  year={2025}
}

@article{bian2025videopainter,
  title={VideoPainter: Any-length Video Inpainting and Editing with Plug-and-Play Context Control},
  author={Bian, Yuxuan and Zhang, Zhaoyang and Ju, Xuan and Cao, Mingdeng and Xie, Liangbin and Shan, Ying and Xu, Qiang},
  journal={arXiv preprint arXiv:2503.05639},
  year={2025}
}

@article{vace,
    title = {VACE: All-in-One Video Creation and Editing},
    author = {Jiang, Zeyinzi and Han, Zhen and Mao, Chaojie and Zhang, Jingfeng and Pan, Yulin and Liu, Yu},
    journal = {arXiv preprint arXiv:2503.07598},
    year = {2025}
}

@article{wan2025wan,
  title={Wan: Open and advanced large-scale video generative models},
  author={Wan, Team and Wang, Ang and Ai, Baole and Wen, Bin and Mao, Chaojie and Xie, Chen-Wei and Chen, Di and Yu, Feiwu and Zhao, Haiming and Yang, Jianxiao and others},
  journal={arXiv preprint arXiv:2503.20314},
  year={2025}
}

@inproceedings{hore2010image,
  title={Image quality metrics: PSNR vs. SSIM},
  author={Hore, Alain and Ziou, Djemel},
  booktitle={ICPR},
  xxxxpages={2366--2369},
  year={2010},
  xxxxorganization={IEEE}
}

@article{wang2004image,
  title={Image quality assessment: from error visibility to structural similarity},
  author={Wang, Zhou and Bovik, Alan C and Sheikh, Hamid R and Simoncelli, Eero P},
  journal=TIP,
  volume={13},
  number={4},
  xxxxpages={600--612},
  year={2004},
  xxxxpublisher={IEEE}
}

@inproceedings{zhang2018unreasonable,
  title={The unreasonable effectiveness of deep features as a perceptual metric},
  author={Zhang, Richard and Isola, Phillip and Efros, Alexei A and Shechtman, Eli and Wang, Oliver},
  booktitle=CVPR,
  xxxxpages={586--595},
  year={2018}
}

@inproceedings{unterthiner2019fvd,
  title={FVD: A new metric for video generation},
  author={Unterthiner, Thomas and Van Steenkiste, Sjoerd and Kurach, Karol and Marinier, Rapha{\"e}l and Michalski, Marcin and Gelly, Sylvain},
  booktitle={ICLR Workshop},
  year={2019}
}

@article{zhao2025objectclear,
  title={ObjectClear: Complete Object Removal via Object-Effect Attention},
  author={Zhao, Jixin and Zhou, Shangchen and Wang, Zhouxia and Yang, Peiqing and Loy, Chen Change},
  journal={arXiv preprint arXiv:2505.22636},
  year={2025}
}

@inproceedings{freeformvideoinpainting,
    author={Chang, Ya-Liang and Liu, Zhe Yu and Lee, Kuan-Ying and Hsu, Winston},
    booktitle={{ICCV}}, 
    title={Free-Form Video Inpainting With 3D Gated Convolution and Temporal PatchGAN}, 
    year={2019}
}

@inproceedings{videoinpaintingbyjointlylearning,
    author = {Wang, Chuan and Huang, Haibin and Han, Xiaoguang and Wang, Jue},
    booktitle = {{AAAI}},
    title = {Video inpainting by jointly learning temporal structure and spatial details},
    year = {2019}
}

@inproceedings{proposalbasedvideocompletion,
    author = {Hu, Yuan-Ting and Wang, Heng and Ballas, Nicolas and Grauman, Kristen and Schwing, Alexander G.},
    title = {Proposal-Based Video Completion},
    booktitle = {{ECCV}},
    year = {2020}
}

@inproceedings{kim2019deepvideoinpainting,
    author = {Dahun Kim and Sanghyun Woo and Joon-Young Lee and In So Kweon},
    booktitle = {{CVPR}},
    title = {Deep Video Inpainting},
    year = {2019},
}

@inproceedings{li2020shortlongterm,
    author = {Ang Li and Shanshan Zhao and Xingjun Ma and Mingming Gong and 
            Jianzhong Qi and Rui Zhang and Dacheng Tao and Ramamohanarao Kotagiri},
    booktitle = {{ECCV}},
    title = {Short-Term and Long-Term Context Aggregation Network for Video Inpainting},
    year = {2020},
}

@inproceedings{zou2021progressive,
    author = {Xueyan Zou and Linjie Yang and Ding Liu and Yong Jae Lee},
    booktitle = {{CVPR}},
    title = {Progressive Temporal Feature Alignment Network for Video Inpainting},
    year = {2021},
}

@inproceedings{lee2019copypaste,
  author = {Sungho Lee and Seoung Wug Oh and DaeYeun Won and Seon Joo Kim},
  booktitle = {{ICCV}},
  title = {Copy-and-Paste Networks for Deep Video Inpainting},
  year = {2019},
}

@inproceedings{cai2022devit,
  author = {Jiayin Cai and Changlin Li and Xin Tao and Chun Yuan and Yu-Wing Tai},
  booktitle = {{ACM MM}},
  title = {DeViT: Deformed Vision Transformers in Video Inpainting},
  year = {2022},
}

@inproceedings{zhou2023propainter,
  title={Propainter: Improving propagation and transformer for video inpainting},
  author={Zhou, Shangchen and Li, Chongyi and Chan, Kelvin CK and Loy, Chen Change},
  booktitle=CVPR,
  xxxxpages={10477--10486},
  year={2023}
}

@inproceedings{ho2020ddpm,
  author    = {Jonathan Ho and Ajay Jain and Pieter Abbeel},
  title     = {Denoising Diffusion Probabilistic Models},
  booktitle = {{NeurIPS}},
  year      = {2020},
}

@inproceedings{shi2024bivdiff,
  author    = {Fengyuan Shi and Jiaxi Gu and Hang Xu and Songcen Xu and Wei Zhang and Limin Wang},
  title     = {Bivdiff: A Training-Free Framework for General-Purpose Video Synthesis via Bridging Image and Video Diffusion Models},
  booktitle = {{CVPR}},
  year      = {2024},
}

@inproceedings{zi2025cococo,
  title={Cococo: Improving text-guided video inpainting for better consistency, controllability and compatibility},
  author={Zi, Bojia and Zhao, Shihao and Qi, Xianbiao and Wang, Jianan and Shi, Yukai and Chen, Qianyu and Liang, Bin and Xiao, Rong and Wong, Kam-Fai and Zhang, Lei},
  booktitle=AAAI,
  xxxxpages={11067--11076},
  year={2025}
}

@inproceedings{ju2024brushnet,
  author = {Xuan Ju and Xian Liu and Xintao Wang and Yuxuan Bian and Ying Shan and Qiang Xu},
  booktitle = {{ECCV}},
  title = {BrushNet: A Plug-and-Play Image Inpainting Model with Decomposed Dual-Branch Diffusion},
  year = {2024},
}

@inproceedings{liu2025erase,
  title={Erase Diffusion: Empowering Object Removal Through Calibrating Diffusion Pathways},
  author={Liu, Yi and Zhou, Hao and Cui, Benlei and Shang, Wenxiang and Lin, Ran},
  booktitle=CVPR,
  xxxxpages={2418--2427},
  year={2025}
}

@article{yu2025omnipaint,
  title={Omnipaint: Mastering object-oriented editing via disentangled insertion-removal inpainting},
  author={Yu, Yongsheng and Zeng, Ziyun and Zheng, Haitian and Luo, Jiebo},
  journal={arXiv preprint arXiv:2503.08677},
  year={2025}
}

@article{wei2025omnieraser,
  title={OmniEraser: Remove Objects and Their Effects in Images with Paired Video-Frame Data},
  author={Wei, Runpu and Yin, Zijin and Zhang, Shuo and Zhou, Lanxiang and Wang, Xueyi and Ban, Chao and Cao, Tianwei and Sun, Hao and He, Zhongjiang and Liang, Kongming and others},
  journal={arXiv preprint arXiv:2501.07397},
  year={2025}
}

@inproceedings{
    miao2025rose,
    title={{ROSE}: Remove Objects with Side Effects in Videos},
    author={Chenxuan Miao and Yutong Feng and Jianshu Zeng and Zixiang Gao and Liu Hantang and Yunfeng Yan and Donglian Qi and Xi Chen and Bin Wang and Hengshuang Zhao},
    booktitle=NeurIPS,
    year={2025},
}

@misc{zi2025minimaxremovertamingbadnoise,
      title={MiniMax-Remover: Taming Bad Noise Helps Video Object Removal}, 
      author={Bojia Zi and Weixuan Peng and Xianbiao Qi and Jianan Wang and Shihao Zhao and Rong Xiao and Kam-Fai Wong},
      year={2025},
      eprint={2505.24873},
      archivePrefix={arXiv},
      primaryClass={cs.CV},
      url={https://arxiv.org/abs/2505.24873}, 
}

@article{fu2026effecterase,
  title={Effecterase: Joint video object removal and insertion for high-quality effect erasing},
  author={Fu, Yang and Zheng, Yike and Dai, Ziyun and Ding, Henghui},
  journal={arXiv preprint arXiv:2603.19224},
  year={2026}
}

@article{chen2024diffusion,
  title={Diffusion forcing: Next-token prediction meets full-sequence diffusion},
  author={Chen, Boyuan and Mart{\'\i} Mons{\'o}, Diego and Du, Yilun and Simchowitz, Max and Tedrake, Russ and Sitzmann, Vincent},
  journal={Advances in Neural Information Processing Systems},
  volume={37},
  pages={24081--24125},
  year={2024}
}

@inproceedings{kodaira2025streamdiffusion,
  title={Streamdiffusion: A pipeline-level solution for real-time interactive generation},
  author={Kodaira, Akio and Xu, Chenfeng and Hazama, Toshiki and Yoshimoto, Takanori and Ohno, Kohei and Mitsuhori, Shogo and Sugano, Soichi and Cho, Hanying and Liu, Zhijian and Tomizuka, Masayoshi and others},
  booktitle={Proceedings of the IEEE/CVF International Conference on Computer Vision},
  pages={12371--12380},
  year={2025}
}

@article{yin2024slow,
  title={From slow bidirectional to fast causal video generators},
  author={Yin, Tianwei and Zhang, Qiang and Zhang, Richard and Freeman, William T and Durand, Fredo and Shechtman, Eli and Huang, Xun},
  journal={arXiv e-prints},
  pages={arXiv--2412},
  year={2024}
}

@article{chen2025skyreels,
  title={Skyreels-v2: Infinite-length film generative model},
  author={Chen, Guibin and Lin, Dixuan and Yang, Jiangping and Lin, Chunze and Zhu, Junchen and Fan, Mingyuan and Zhang, Hao and Chen, Sheng and Chen, Zheng and Ma, Chengcheng and others},
  journal={arXiv preprint arXiv:2504.13074},
  year={2025}
}

@article{teng2025magi,
  title={MAGI-1: Autoregressive Video Generation at Scale},
  author={Teng, Hansi and Jia, Hongyu and Sun, Lei and Li, Lingzhi and Li, Maolin and Tang, Mingqiu and Han, Shuai and Zhang, Tianning and Zhang, WQ and Luo, Weifeng and others},
  journal={arXiv preprint arXiv:2505.13211},
  year={2025}
}

@article{guo2025long,
  title={Long context tuning for video generation},
  author={Guo, Yuwei and Yang, Ceyuan and Yang, Ziyan and Ma, Zhibei and Lin, Zhijie and Yang, Zhenheng and Lin, Dahua and Jiang, Lu},
  journal={arXiv preprint arXiv:2503.10589},
  year={2025}
}

@article{lin2025autoregressive,
  title={Autoregressive Adversarial Post-Training for Real-Time Interactive Video Generation},
  author={Lin, Shanchuan and Yang, Ceyuan and He, Hao and Jiang, Jianwen and Ren, Yuxi and Xia, Xin and Zhao, Yang and Xiao, Xuefeng and Jiang, Lu},
  journal={arXiv preprint arXiv:2506.09350},
  year={2025}
}

@article{huang2026self,
  title={Self forcing: Bridging the train-test gap in autoregressive video diffusion},
  author={Huang, Xun and Li, Zhengqi and He, Guande and Zhou, Mingyuan and Shechtman, Eli},
  journal={Advances in Neural Information Processing Systems},
  volume={38},
  pages={167283--167308},
  year={2026}
}

@misc{song2023consistencymodels,
      title={Consistency Models}, 
      author={Yang Song and Prafulla Dhariwal and Mark Chen and Ilya Sutskever},
      year={2023},
      eprint={2303.01469},
      archivePrefix={arXiv},
      primaryClass={cs.LG},
      url={https://arxiv.org/abs/2303.01469}, 
}

@article{wang2023videolcm,
  title={Videolcm: Video latent consistency model},
  author={Wang, Xiang and Zhang, Shiwei and Zhang, Han and Liu, Yu and Zhang, Yingya and Gao, Changxin and Sang, Nong},
  journal={arXiv preprint arXiv:2312.09109},
  year={2023}
}

@article{salimans2022progressive,
  title={Progressive distillation for fast sampling of diffusion models},
  author={Salimans, Tim and Ho, Jonathan},
  journal={arXiv preprint arXiv:2202.00512},
  year={2022}
}

@article{song2023improved,
  title={Improved techniques for training consistency models},
  author={Song, Yang and Dhariwal, Prafulla},
  journal={arXiv preprint arXiv:2310.14189},
  year={2023}
}

@inproceedings{liu2023instaflow,
  title={Instaflow: One step is enough for high-quality diffusion-based text-to-image generation},
  author={Liu, Xingchao and Zhang, Xiwen and Ma, Jianzhu and Peng, Jian and others},
  booktitle={The Twelfth International Conference on Learning Representations},
  year={2023}
}

@InProceedings{Dong_2025_CVPR,
    author    = {Dong, Linwei and Fan, Qingnan and Guo, Yihong and Wang, Zhonghao and Zhang, Qi and Chen, Jinwei and Luo, Yawei and Zou, Changqing},
    title     = {TSD-SR: One-Step Diffusion with Target Score Distillation for Real-World Image Super-Resolution},
    booktitle = {Proceedings of the IEEE/CVF Conference on Computer Vision and Pattern Recognition (CVPR)},
    month     = {June},
    year      = {2025},
    pages     = {23174-23184}
}

@inproceedings{yin2024dmd,
  title     = {One-step Diffusion with Distribution Matching Distillation},
  author    = {Yin, Tianwei and Gharbi, Micha{\"e}l and Zhang, Richard and Shechtman, Eli and Durand, Fr{\'e}do and Freeman, William T. and Park, Taesung},
  booktitle = {Proceedings of the IEEE/CVF Conference on Computer Vision and Pattern Recognition (CVPR)},
  pages     = {6613--6623},
  year      = {2024},
  month     = {June},
  doi       = {10.1109/CVPR52733.2024.00632},
  url       = {https://openaccess.thecvf.com/content/CVPR2024/html/Yin_One-step_Diffusion_with_Distribution_Matching_Distillation_CVPR_2024_paper.html}
}

@inproceedings{yin2024dmd2,
  title     = {Improved Distribution Matching Distillation for Fast Image Synthesis},
  author    = {Yin, Tianwei and Gharbi, Micha{\"e}l and Park, Taesung and Zhang, Richard and Shechtman, Eli and Durand, Fr{\'e}do and Freeman, William T.},
  booktitle = {Advances in Neural Information Processing Systems},
  volume    = {37},
  pages     = {47455--47487},
  year      = {2024},
  editor    = {Globerson, A. and Mackey, L. and Belgrave, D. and Fan, A. and Paquet, U. and Tomczak, J. and Zhang, C.},
  publisher = {Curran Associates, Inc.},
  doi       = {10.52202/079017-1505},
  url       = {https://proceedings.neurips.cc/paper_files/paper/2024/file/54dcf25318f9de5a7a01f0a4125c541e-Paper-Conference.pdf}
}

@article{lin2024sdxl,
  title={Sdxl-lightning: Progressive adversarial diffusion distillation},
  author={Lin, Shanchuan and Wang, Anran and Yang, Xiao},
  journal={arXiv preprint arXiv:2402.13929},
  year={2024}
}

@article{lin2025diffusion,
  title={Diffusion adversarial post-training for one-step video generation},
  author={Lin, Shanchuan and Xia, Xin and Ren, Yuxi and Yang, Ceyuan and Xiao, Xuefeng and Jiang, Lu},
  journal={arXiv preprint arXiv:2501.08316},
  year={2025}
}

@article{wang2025seedvr2,
  title={Seedvr2: One-step video restoration via diffusion adversarial post-training},
  author={Wang, Jianyi and Lin, Shanchuan and Lin, Zhijie and Ren, Yuxi and Wei, Meng and Yue, Zongsheng and Zhou, Shangchen and Chen, Hao and Zhao, Yang and Yang, Ceyuan and others},
  journal={arXiv preprint arXiv:2506.05301},
  year={2025}
}

@misc{chen2025doveefficientonestepdiffusion,
      title={DOVE: Efficient One-Step Diffusion Model for Real-World Video Super-Resolution}, 
      author={Zheng Chen and Zichen Zou and Kewei Zhang and Xiongfei Su and Xin Yuan and Yong Guo and Yulun Zhang},
      year={2025},
      eprint={2505.16239},
      archivePrefix={arXiv},
      primaryClass={cs.CV},
      url={https://arxiv.org/abs/2505.16239}, 
}

@article{zhuang2025flashvsr,
  title={Flashvsr: Towards real-time diffusion-based streaming video super-resolution},
  author={Zhuang, Junhao and Guo, Shi and Cai, Xin and Li, Xiaohui and Liu, Yihao and Yuan, Chun and Xue, Tianfan},
  journal={arXiv preprint arXiv:2510.12747},
  year={2025}
}

@article{li2025personalive,
  title={PersonaLive! Expressive Portrait Image Animation for Live Streaming},
  author={Li, Zhiyuan and Pun, Chi-Man and Fang, Chen and Wang, Jue and Cun, Xiaodong},
  journal={arXiv preprint arXiv:2512.11253},
  year={2025}
}

@article{kingma2013auto,
  title={Auto-encoding variational bayes},
  author={Kingma, Diederik P and Welling, Max},
  journal={arXiv preprint arXiv:1312.6114},
  year={2013}
}

@inproceedings{esser2024scaling,
  title={Scaling rectified flow transformers for high-resolution image synthesis},
  author={Esser, Patrick and Kulal, Sumith and Blattmann, Andreas and Entezari, Rahim and M{\"u}ller, Jonas and Saini, Harry and Levi, Yam and Lorenz, Dominik and Sauer, Axel and Boesel, Frederic and others},
  booktitle=ICML,
  year={2024}
}

@article{raffel2020exploring,
  title={Exploring the limits of transfer learning with a unified text-to-text transformer},
  author={Raffel, Colin and Shazeer, Noam and Roberts, Adam and Lee, Katherine and Narang, Sharan and Matena, Michael and Zhou, Yanqi and Li, Wei and Liu, Peter J},
  journal=JMLR,
  volume={21},
  number={140},
  xxxxpages={1--67},
  year={2020}
}

@inproceedings{radford2021learning,
  title={Learning transferable visual models from natural language supervision},
  author={Radford, Alec and Kim, Jong Wook and Hallacy, Chris and Ramesh, Aditya and Goh, Gabriel and Agarwal, Sandhini and Sastry, Girish and Askell, Amanda and Mishkin, Pamela and Clark, Jack and others},
  booktitle=ICML,
  xxxxpages={8748--8763},
  year={2021},
  xxxxorganization={PmLR}
}

@inproceedings{vaswani2017attention,
  title={Attention is all you need},
  author={Vaswani, Ashish and Shazeer, Noam and Parmar, Niki and Uszkoreit, Jakob and Jones, Llion and Gomez, Aidan N and Kaiser, {\L}ukasz and Polosukhin, Illia},
  booktitle=NeurIPS,
  year={2017}
}

@article{zhu2026causal,
  title={Causal Forcing: Autoregressive Diffusion Distillation Done Right for High-Quality Real-Time Interactive Video Generation},
  author={Zhu, Hongzhou and Zhao, Min and He, Guande and Su, Hang and Li, Chongxuan and Zhu, Jun},
  journal={arXiv preprint arXiv:2602.02214},
  year={2026}
}

@InProceedings{huang2023vbench,
     title={{VBench}: Comprehensive Benchmark Suite for Video Generative Models},
     author={Huang, Ziqi and He, Yinan and Yu, Jiashuo and Zhang, Fan and Si, Chenyang and Jiang, Yuming and Zhang, Yuanhan and Wu, Tianxing and Jin, Qingyang and Chanpaisit, Nattapol and Wang, Yaohui and Chen, Xinyuan and Wang, Limin and Lin, Dahua and Qiao, Yu and Liu, Ziwei},
     booktitle={{CVPR}},
     year={2024}
 }

@article{ba2016layer,
  title={Layer normalization},
  author={Ba, Jimmy Lei and Kiros, Jamie Ryan and Hinton, Geoffrey E},
  journal={arXiv preprint arXiv:1607.06450},
  year={2016}
}

@article{hendrycks2016gaussian,
  title={Gaussian Error Linear Units (Gelus)},
  author={Hendrycks, D},
  journal={arXiv preprint arXiv:1606.08415},
  year={2016}
}

@misc{shin2026motionstreamrealtimevideogeneration,
      title={MotionStream: Real-Time Video Generation with Interactive Motion Controls}, 
      author={Joonghyuk Shin and Zhengqi Li and Richard Zhang and Jun-Yan Zhu and Jaesik Park and Eli Shechtman and Xun Huang},
      year={2026},
      eprint={2511.01266},
      archivePrefix={arXiv},
      primaryClass={cs.CV},
      url={https://arxiv.org/abs/2511.01266}, 
}

@article{xiao2023efficient,
  title={Efficient streaming language models with attention sinks},
  author={Xiao, Guangxuan and Tian, Yuandong and Chen, Beidi and Han, Song and Lewis, Mike},
  journal={arXiv preprint arXiv:2309.17453},
  year={2023}
}
